\documentclass[sigconf]{acmart}
\makeatletter
\@ifundefined{Bbbk}{}{}
\makeatother

\usepackage{amsmath}
\usepackage{amssymb}
\usepackage{amsfonts}
\usepackage{makecell}
\usepackage{multirow}      
\usepackage{color}         
\usepackage{subcaption}    
\usepackage[ruled,vlined]{algorithm2e}  

\setcopyright{acmcopyright}

\DeclareUnicodeCharacter{2212}{\ensuremath{-}}

\acmConference[ICVGIP'25]{16th Indian Conference on Computer Vision, Graphics and Image Processing}{December 2025}{Himachal Pradesh, India}
\acmYear{2025}
\copyrightyear{2025}

\acmPrice{15.00}

\begin{document}
\title{Compressive Modeling and Visualization of Multivariate Scientific Data using Implicit Neural Representation}


\author{Abhay Kumar Dwivedi, Shanu Saklani, and Soumya Dutta}
\email{abhaykd@cse.iitk.ac.in, shanu@cse.iitk.ac.in, soumyad@cse.iitk.ac.in}
\affiliation{%
  \institution{Indian Institute of Technology Kanpur (IIT Kanpur)}
  \city{Kanpur}
  \state{Uttar Pradesh}
  \country{India}
  \postcode{208016}
}
\renewcommand{\shortauthors}{A. Dwivedi, S. Saklani, and S. Dutta}

\begin{abstract}
The extensive adoption of Deep Neural Networks has led to their increased utilization in challenging scientific visualization tasks. Recent advancements in building compressed data models using implicit neural representations have shown promising results for tasks like spatiotemporal volume visualization and super-resolution. Inspired by these successes, we develop compressed neural representations for multivariate datasets containing tens to hundreds of variables. Our approach utilizes a single network to learn representations for all data variables simultaneously through parameter sharing. This allows us to achieve state-of-the-art data compression. Through comprehensive evaluations, we demonstrate superior performance in terms of reconstructed data quality, rendering and visualization quality, preservation of dependency information among variables, and storage efficiency.
\end{abstract}

%
%



\maketitle

\section{Introduction}

Efficiently analyzing and visualizing multivariate scientific datasets is essential for gaining insights into various natural phenomena. Scientists from diverse domains utilize computational simulations to generate complex multivariate datasets, often comprising hundreds of variables. Instead of examining each variable independently, exploring the interaction among multiple variables concurrently is a powerful approach for studying intricate scientific phenomena~\cite{codda}. For example, in weather simulation, while studying hurricanes, the interaction between pressure, wind velocity, and precipitation is often jointly examined to enable a comprehensive understanding of the complex interplay between these variables.

As the size, complexity, and number of variables in multivariate datasets increase, scientists across various domains face significant challenges in efficiently storing, managing, analyzing, and visualizing such data. This challenge becomes particularly pronounced when dealing with datasets containing hundreds of variables, such as those in climate and weather forecasting simulations. Consequently, data scientists are exploring various compression methods~\cite{codda, TTHRESH, pmi_sampling}. Recently, there has been considerable interest in using deep learning models to represent univariate data compactly~\cite {levine_neural_compression, coordnet}. Although these methods have achieved impressive results, they have yet to address the direct compression of multivariate datasets. Moreover, it is crucial to comprehensively evaluate such models to ensure: (1) faithful on-demand reconstruction of all variables with high accuracy, (2) preservation of variable relationships (dependencies) in reconstructed data, and (3) meaningful results in various types of multivariate analysis tasks.

In this work, the primary focus is on thoroughly studying the viability of implicit neural representations (INRs) for modeling a large number of variables (tens to hundreds) jointly. We investigate whether using a single INR to model all variables jointly can efficiently learn such a diverse set of variables. We aim to evaluate the achievable reconstruction quality of this approach and compare it with alternative data reduction methods. To achieve this, we employ a residual SIREN architecture. By using a single network, we leverage parameter sharing across all data variables, enabling us to achieve a high compression ratio. We conduct a rigorous evaluation of the network's prediction quality from different perspectives, comparing it with other competitive approaches to demonstrate its superiority. In addition to comparing prediction quality using conventional data reconstruction and image-space metrics, we also perform a detailed comparative analysis using important multivariate application-specific tasks. We examine the accuracy of inter-variable relationships and assess their performance in conducting multivariate analyses. Finally, we investigate the performance of our model under varying conditions, including changes in the number of variables, variations in the amount of training data, and adjustments to the network depth. Our findings suggest that implicit neural representations are effective in capturing intricate patterns from multivariate scientific datasets, which contain numerous variables, and can efficiently represent complex multivariate data in a compact manner.

\section{Related Works}
\textbf{Multivariate Data Visualization.}
Multivariate data analysis is a popular research area with many applications in various domains~\cite{de2003visual, wong199430}. Sauber et al.~\cite{corr_vis_1} propose a multifield-graph-based technique for determining the local correlation coefficients between various variables, facilitating the visualization and analysis of multivariate data. Gosink et al.~\cite{gosink2010application} use local statistical distributions to improve it further for Query-driven analysis. To identify the interesting regions of multivariate datasets, local statistical complexities are utilized by Jänicke et al.~\cite{janicke2007multifield}. For time-varying multivariate datasets, information theory is used to discover the causal relationship among variables~\cite{wang2011analyzing}. Biswas et al.~\cite{vis13_info_theory} use information theory to study variable interaction, and Dutta et al.~\cite{pmi_sampling} perform multivariate sampling.

\textbf{Deep Learning for Scientific Visualization.}
Deep learning has found numerous applications in scientific visualization.
Techniques for generating compact neural representations of scientific scalar data are proposed by Lu et al.~\cite{lu2021compressive} and Weiss et al.~\cite{weiss2022fast}. Hong et al.~\cite{hong2019dnn}, He et al.\cite{he2019insitunet}, and Berger et al.~\cite{hong2019dnn} study the visualization of scalar field data using volume-rendered images. Weiss et al.~\cite{weiss2019volumetric} use isosurfaces for the same and further explored an adaptive sampling guided method for volume data visualization~\cite{weiss2020learning}. 
For compressing the volume data, new models for domain knowledge-aware latent space generation techniques for scalar data are also proposed~\cite{shen2022idlat}.
Generation of high-resolution spatiotemporal volumes from low-resolution data~\cite{
 TSR-TVD, wurster2022deep, SSRTVD} is another research area of focus.
Also, for visualizing and exploring the parameter spaces for ensemble data, DNNs are utilized as surrogates~\cite{he2019insitunet, shi2022gnn, shi2022vdl}.
GCNs are explored by Han and Wang~\cite{han2022surfnet} for learning surface representations. Han et al.~\cite{han2020v2v} suggest a variable-to-variable translation technique for scientific multivariate data.  There are many more applications of deep learning in scientific visualization, and for a more detailed review, please refer to the state-of-the-art survey~\cite{wang2022dl4scivis}.

\textbf{Data Compression and Summarization.}
Data compression and summarization techniques have become increasingly pivotal as scientific datasets have grown significantly in recent years. Scientists have adopted several approaches to reduce the size of data with minimal loss of information. One such approach is to use compression techniques~\cite{zfp, sz, TTHRESH} to rapidly reduce the size of datasets and then decompress them during analysis and visualization. Woodring et al.~\cite{woodring_jpeg} apply Wavelet-based compression to compress climate data. Statistical methods for data summarization and reduction have been extensively studied. More importantly, statistical  Copula-model-based~\cite{codda} and multivariate sampling~\cite{pmi_sampling} approaches have been investigated recently for summarizing multivariate datasets. For a detailed summary of existing data reduction techniques, please refer to~\cite{data_reduction_survey_sam}.

\section{Learning Multivariate Data using Implicit Neural Representation}

\subsection{Implicit Neural Representation}
INRs with periodic activation functions have emerged as a promising approach for learning coordinate-based datasets such as images, scientific data, shape-based data, etc. Sitzmann et al.~\cite{siren} showed that a feed-forward neural network with sinusoidal activation function, when initialized with a carefully chosen weighting scheme, can be used to model various types of coordinate-based data efficiently. Such networks are known as SIREN (sinusoidal representation network)~\cite{siren}. Several variations of SIREN have been employed in the visualization research community to learn scalar field data~\cite{levine_neural_compression}, and model representations of various visualization types~\cite{coordnet}. In a recent research, Tang and Wang used implicit neural networks to generate spatiotemporal super-resolution of scientific data from low-resolution representations~\cite{stsrinr}. 

\subsection{Motivation}
In this work, we employ a modified SIREN for modeling multivariate datasets with a large number of variables. In multivariate data, different variables contain different features and patterns, adding to the complexity of the data. Subsets of variables also often show the existence of complex local and global dependencies~\cite{vis13_info_theory}. Hence, the primary motivation of our work is to investigate whether a SIREN-based neural network can accurately learn complex variable dynamics with a very large number of variables in the dataset. A potential reason for reduced accuracy on this task could be that the data variability across a large number of variables negatively impacts training, causing performance deterioration. Such a network may also require many hidden layers to successfully learn the multivariate data patterns, increasing the model complexity. Besides these possible pitfalls, we also aim to investigate the amount of compression that can be achieved by using such a neural representation compared to existing state-of-the-art data reduction techniques, such as tensor compression~\cite{TTHRESH}. Finally, we also seek to evaluate (a) how accurately such a network preserves the dependency among the variables and (b) the effectiveness of the reconstructed data in performing important multivariate analysis tasks.  

\begin{figure}[tb]
\centering
\includegraphics[width=\linewidth]{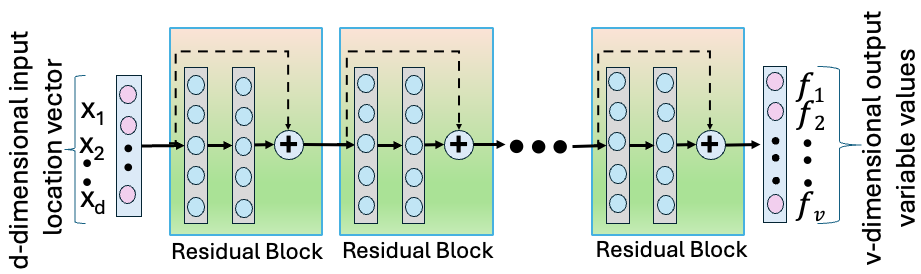}
\caption{Schematic architecture of MVNet. It uses a residual SIREN architecture to enhance its learning capability.}
\label{model_arch}
\end{figure}

\subsection{Model Architecture}
We build our \textbf{M}ulti\textbf{V}ariate data representation \textbf{Net}work (MVNet), based on a SIREN architecture~\cite{siren}. Our goal is to learn the representation of multivariate data as a function, which is represented by the parameters of a neural network. We want the size of the neural network to be significantly smaller than the size of the raw multivariate data so that a compressed multivariate data model is produced for efficient analysis and visualization. Our model represents a function that takes a $d$ dimensional location vector, i.e., for 3D volumetric data, $d=3$, and we pass the coordinate of the grid point as input and predict a $v$ dimensional vector output where $v$ is the number of variables in the multivariate dataset. Essentially, our neural network learns a function $\mathcal{F(\theta)}:\mathbb{R}^d \mapsto \mathbb{R}^v$, where $\theta$ represents the parameters of the neural network. 

We build a multilayer perceptron consisting of $d$ neurons as inputs, followed by $l$ hidden layers, and an output layer containing $v$ neurons. For boosting the model's learning capability and training a deep network stably, we enhance this architecture by incorporating residual blocks and skip connections~\cite{resnet}. By formulating the learning of $v$ variables using a single neural network, we pose this as a simple multi-task learning problem~\cite{multitask_survey}. As the variables in the scientific datasets are often correlated~\cite{vis13_info_theory, codda}, we believe that such inter-dependencies can help our network to learn the diverse patterns of all the variables efficiently, where learning representations of one variable is considered as a single task, and the network learns all the variable representations jointly. The applicability of multi-task learning where the dataset has several related variables has been shown to be effective by researchers in the past~\cite{multitask_survey, multitask1, multitask2}. Furthermore, using a single network to learn all the variables jointly for a multivariate dataset also helps us to obtain a compact representation of the entire data since the network parameters are shared, and hence the network size is minimal.

In Fig.~\ref{model_arch}, a schematic of our model architecture is shown. We use a sinusoidal activation function for its superior performance over alternative activation functions such as ReLU and Sigmoid~\cite{siren}. It can be seen that the network takes a $d$ dimensional location vector as input and predicts a $v$ dimensional vector output, where each element of the output vector indicates the value of an individual variable. To train this network, we use the Adam optimizer~\cite{kiba14} and the conventional mean squared error (MSE) loss ($\mathcal{L}_{mse}$) function. We conduct a thorough hyperparameter search to find the best parameters and learning rate for the Adam optimizer. While computing the MSE loss, we use equal weight on loss values for all the variables. To train MVNet in a stable manner, we normalize the input coordinates between 0 and 1 and rescale the values of each variable between -1 and 1.  

\section{Comparative Study}

To show the efficacy of MVNet over the existing data compression and summarization approaches, we present a thorough comparative study. We use four multivariate datasets to demonstrate the results. The dimensionality, number of variables, and spatial resolution of the datasets are reported in Table~\ref{datadesc_table}. We use a GPU server with NVIDIA GeForce GTX $1080$Ti GPUs with $12$GB GPU memory for all the experimentation. All the models are implemented in PyTorch~\cite{pytorch2019}. Hurricane Isabel data was produced by the Weather Research and Forecast model, courtesy of NCAR and the U.S. National Science Foundation. The Turbulent Combustion dataset is made available by Dr. Jacqueline Chen at Sandia Laboratories through the U.S. Department of Energy's SciDAC Institute for Ultrascale Visualization. The Climate50 and Climate100 datasets were obtained from the Earth System Grid Federation's online data repository. The dataset was generated by Energy Exascale Earth System Model (E3SM).
\begin{table}[thb]
\centering
\caption{Description of datasets used in the experimentation.}
\label{datadesc_table}
\resizebox{0.8\linewidth}{!}{
\begin{tabular}{|c|c|c|c|}
\hline
\textbf{Dataset}      & \textbf{Dimensionality} & \textbf{\#Variables} & \textbf{Resolution} \\ \hline
\textbf{Combustion} & 3D                      & 5                    & 240$\times$360$\times$60          \\ \hline
\textbf{Isabel  }   & 3D                      & 14                   & 250$\times$250$\times$50          \\ \hline
\textbf{Climate50 }             & 2D                      & 53                   & 2880$\times$1440           \\ \hline
\textbf{Climate100  }           & 2D                      & 100                  & 2880$\times$1440           \\ \hline
\end{tabular}
}
\end{table}
First we compare MVNet with three different data compression approaches. The first one is the baseline, where the multivariate dataset is first sub-sampled into a lower resolution grid using linear interpolation, and then again linear interpolation is used to up-sample the data back to the original grid resolution. Typically such sub-sampling is often used as a means of data reduction. We call this method as \textit{LERP}. The second method is the state-of-the-art tensor compression, \textit{TTHRESH}~\cite{TTHRESH}. The third method is another widely used floating point compression method \textit{Zfp}~\cite{zfp}. We compare the reconstruction quality of all the variables using Peak Signal to Noise Ratio (PSNR). To conduct a comparison in feature space, we use isocontour similarity to estimate the accuracy of isocontours generated by each method. Next, to evaluate the visualization quality in image space, we render images of data variables and compare their accuracy by comparing them against the ground truth. Finally, we compare MVNet with the statistical Copula-based summarization method proposed in~\cite{codda}.

\subsection{Comparison of Reconstructed Data Quality}
\begin{figure}[tb]
\centering
\includegraphics[width=0.85\linewidth, height=1.8in]{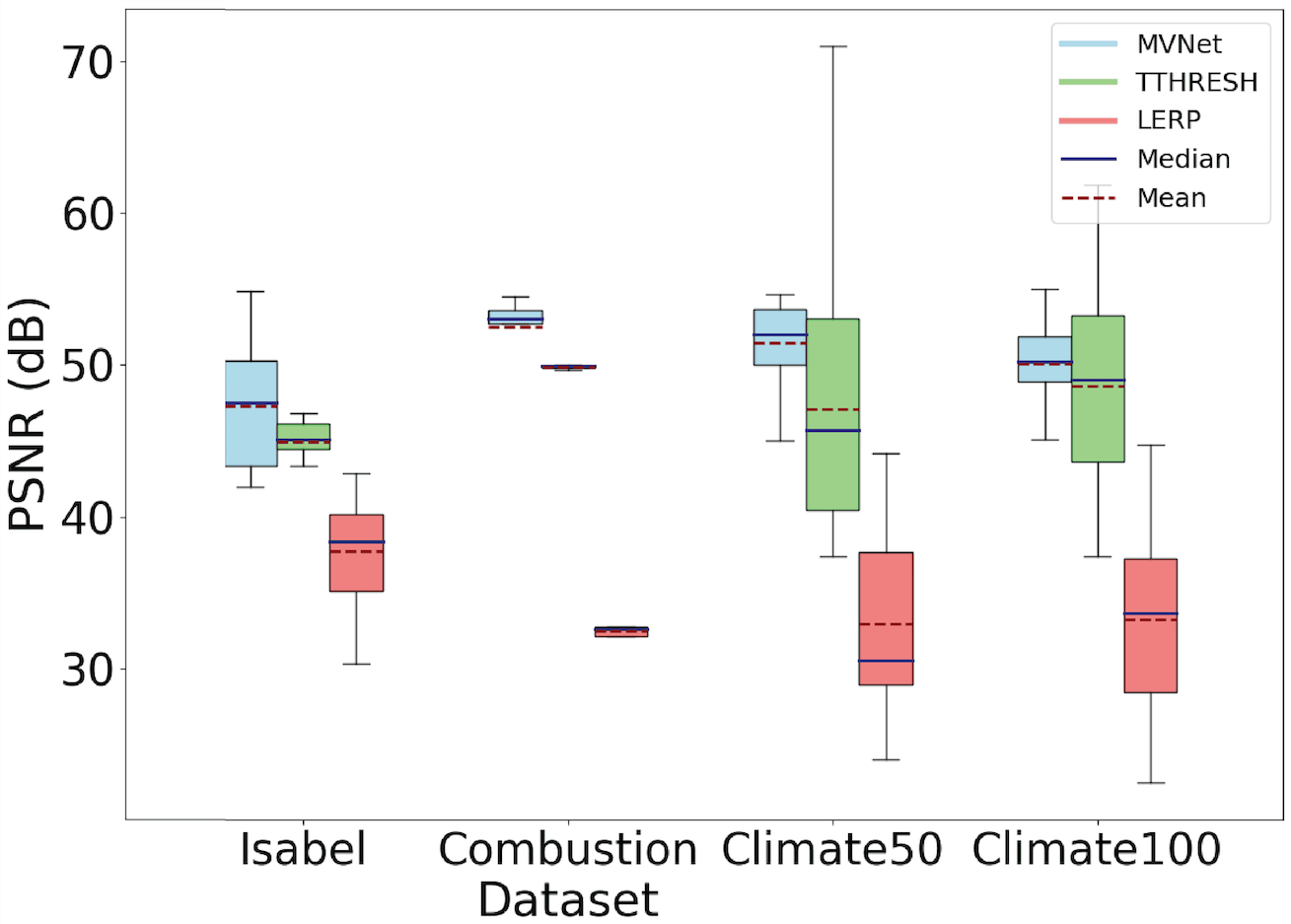}
\caption{Comparison of reconstruction quality for MVNet, LERP, and TTHRESH. We provide a boxplot for each method for each dataset. We observe that MVNet consistently delivers superior PSNR compared to LERP and TTHRESH.}
\label{psnrboxplot}
\end{figure}

The quality of the reconstructed data is quantitatively compared by computing the PSNR for each data variable for each method. Our goal is to compare MVNet against TTHRESH, Zfp, and LERP by comparing the reconstructed data quality given comparable storage footprint. We find that MVNet with $10$ residual blocks with $120$ neurons in each layer produces the best quality vs. storage trade-off. In Table~\ref{psnr_table}, the results for all the four methods are provided. We use Zfp with both fixed bitrate (FBR) and fixed absolute error (FAE) as compression guideline. Since the datasets contain a large number of variables and due to space constraints, we present the average PNSR of all the variables in Table~\ref{psnr_table}. We observe that TTHRESH, Zfp, and LERP methods produce reconstructed data with lower PSNR than MVNet, even with slightly higher storage. Hence, MVNet gives the best trade-off between storage and PSNR. We further observe that, between Zfp and TTHRESH,  TTHRESH gives better trade-off. Hence, we use only TTHRESH for the subsequent analyses and visualization. Next, Fig~\ref{psnrboxplot} shows a joint boxplot comparing PSNR for MVNet, LERP, and TTHRESH across various datasets. In each boxplot, we show the mean, median and further map the maximum and minimum variable PSNR values to the whiskers. A boxplot is drawn for each dataset for each method. The boxplots illustrate that MVNet achieves higher PSNR, compared to LERP and TTHRESH, showing its superior and consistent reconstruction performance. Finally, in Table~\ref{error_stats_tab}, we report the average absolute maximum error, 95th percentile absolute error and the fraction above the tolerance level ($0.05$) for all the variables for MVNet. Note that, we compute these error statistics on variable values scaled between -1 to 1 to obtain consistent and comparable error statistics.  

\begin{table}[thb]
\centering
\caption{Quantitative comparison of average reconstruction quality for all the variables vs. storage overhead for MVNet and other methods. All raw datasets are stored using VTK data format. It is observed that MVNet produces higher reconstruction quality with the minimum storage.}
\label{psnr_table}
\renewcommand{\arraystretch}{1.2}
\resizebox{\linewidth}{!}{
\begin{tabular}{|c|c|cc|cc|cc|cc|cc|}
\hline
\textbf{Dataset} & 
\begin{tabular}[c]{@{}c@{}}Raw \\ Stor.\\ (MB)\end{tabular} & 
\multicolumn{2}{c|}{\textbf{MVNet}} & 
\multicolumn{2}{c|}{\textbf{TTHRESH}} & 
\multicolumn{2}{c|}{\textbf{LERP}} &
\multicolumn{2}{c|}{\textbf{ZFP (FBR)}} &
\multicolumn{2}{c|}{\textbf{ZFP (FAE)}} \\
\cline{3-12}
& & 
\begin{tabular}[c]{@{}c@{}}Stor.\\ (KB)\end{tabular} & 
\begin{tabular}[c]{@{}c@{}}PSNR\\ (dB)\end{tabular} & 
\begin{tabular}[c]{@{}c@{}}Stor.\\ (KB)\end{tabular} & 
\begin{tabular}[c]{@{}c@{}}PSNR\\ (dB)\end{tabular} & 
\begin{tabular}[c]{@{}c@{}}Stor.\\ (KB)\end{tabular} & 
\begin{tabular}[c]{@{}c@{}}PSNR\\ (dB)\end{tabular} &
\begin{tabular}[c]{@{}c@{}}Stor.\\ (KB)\end{tabular} & 
\begin{tabular}[c]{@{}c@{}}PSNR\\ (dB)\end{tabular} & 
\begin{tabular}[c]{@{}c@{}}Stor.\\ (KB)\end{tabular} & 
\begin{tabular}[c]{@{}c@{}}PSNR\\ (dB)\end{tabular} \\
\hline
Comb. & 108 & 1160 & 52.51 & 1608 & 49.89 & 1816 & 32.47 & 1400 & 30.45 & 4284 & 46.88\\
Isabel & 158 & 1164 & 47.32 & 2072 & 44.94 & 2164 & 37.74 & 1960 & 34.87 & 1856 & 39.07\\
CLM50  & 842 & 1184 & 51.27 & 2792 & 47.10 & 2052 & 33.01 & 21836 & 18.79 & 8808 & 17.97\\
CLM100 & 1568 & 1208 & 49.68 & 4568 & 48.62 & 3076 & 33.25 & 47200 & 23.99 & 9160 & 22.2\\
\hline
\end{tabular}}
\end{table}

\begin{table}[thb]
\caption{Maximum absolute error, 95th percentile absolute error, and fraction above tolerance (0.05) for MVNet for all the datasets.}
\label{error_stats_tab}
\resizebox{\linewidth}{!}{%
\begin{tabular}{|c|c|c|c|}
\hline
\textbf{DataSet} & \textbf{Max. Error} & \textbf{95th Percentile error} & \textbf{Frac. above tolerance (0.05)} \\ \hline
\textbf{Combustion} & 0.056 & 0.012 & 0.000012 \\ \hline
\textbf{Isabel}     & 0.371 & 0.026 & 0.004    \\ \hline
\textbf{climate50}  & 0.103 & 0.021 & 0.013    \\ \hline
\textbf{climate100} & 0.144 & 0.029 & 0.06     \\ \hline
\end{tabular}%
}
\end{table}

\begin{figure}[!thb]
\centering
\begin{subfigure}[t]{0.23\linewidth}
    \centering
    \includegraphics[width=\linewidth]{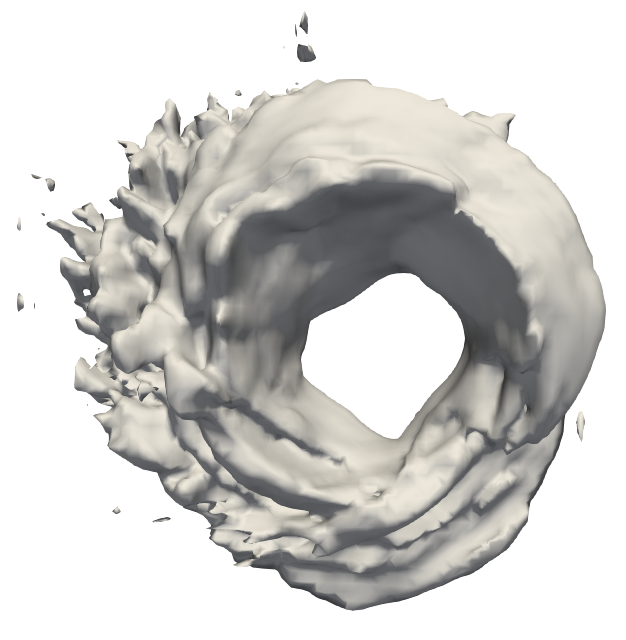}
    \caption{GT}
    \label{isabel_isosurf_41_GT}
\end{subfigure}
~
\begin{subfigure}[t]{0.23\linewidth}
    \centering
    \includegraphics[width=\linewidth]{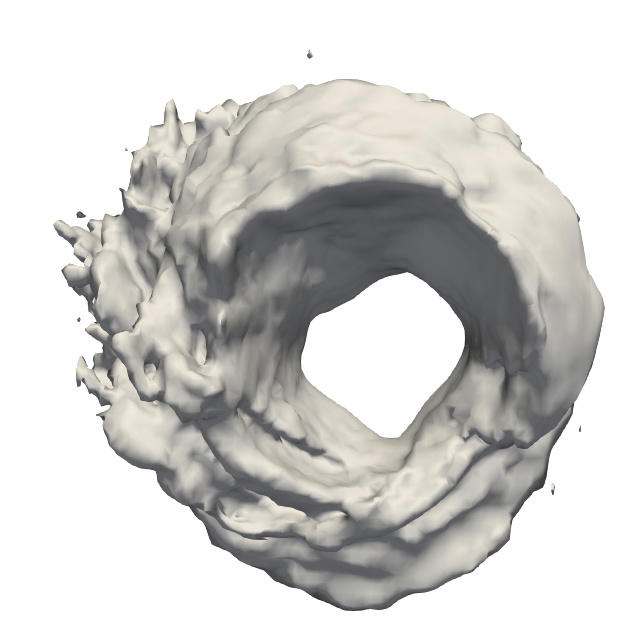}
    \caption{MVNet}
    \label{isabel_isosurf_41_mvnet}
\end{subfigure}
~
\begin{subfigure}[t]{0.23\linewidth}
    \centering
    \includegraphics[width=\linewidth]{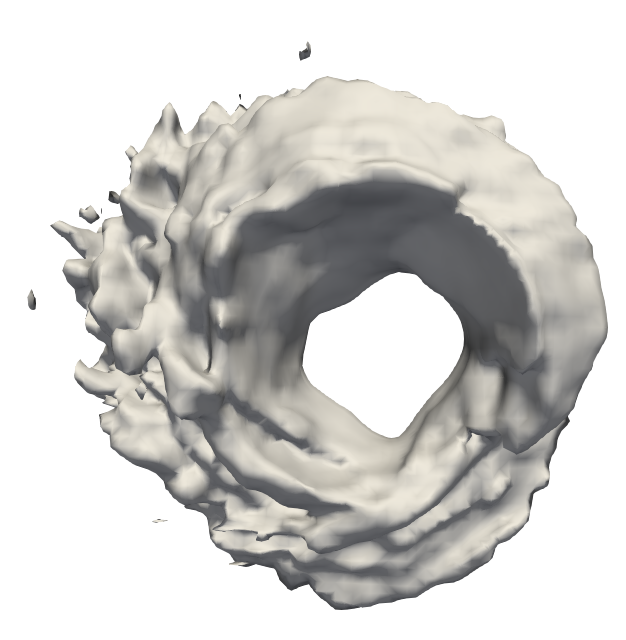}
      \caption{TTHRESH}
    \label{isabel_isosurf_41_tthresh}
\end{subfigure}
~
\begin{subfigure}[t]{0.23\linewidth}
    \centering
    \includegraphics[width=\linewidth]{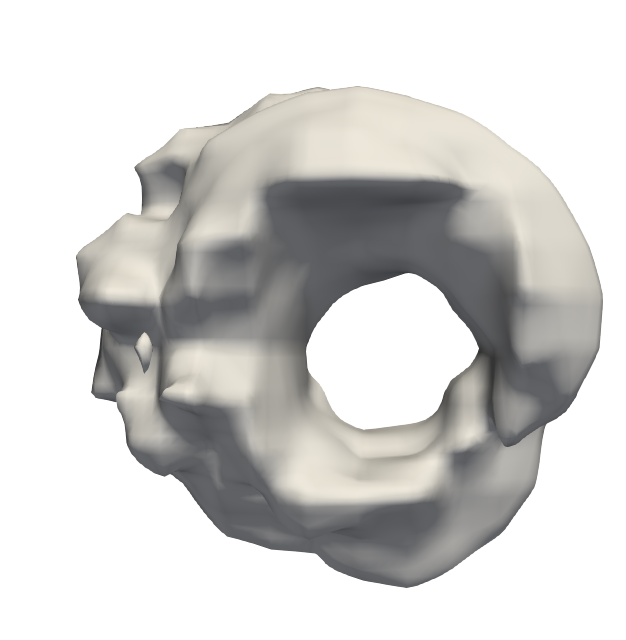}
    \caption{LERP}
    \label{isabel_isosurf_41_lerp}
\end{subfigure}
\caption{Visualization of isosurface of Velocity Magnitude at isovalue=41 of Isabel dataset with ground truth (GT) shown in Fig.~\ref{isabel_isosurf_41_GT}. Isosurface produced by LERP is least accurate while MVNet and TTHRESH produce visually comparable results.}
\label{isabel_isosurf_vis}
\end{figure}

\subsection{Comparison of Reconstructed Features}
Isocontours are considered important feature representations in scientific datasets. Hence, we evaluate the accuracy of the extracted isocontours for the three methods. Given a dataset, we first uniformly randomly select $20$ isovalues for each variable and then extract those isosurfaces. Then, we compute the Hausdorff distance and Chamfer distance~\cite{champher} for each isocontour against the ground truth isocontours extracted from the raw data. This study is conducted for the 3D datasets. The average isocontour distance over 20 isocontours is computed for each variable, and finally, the average of such variable-wise average values is computed for each dataset. Table~\ref{cf_table} shows the results of this feature-based comparison. We observe that, on average, the accuracy of the isocontours is the highest for MVNet compared to TTHRESH and LERP. In Fig.~\ref{isabel_isosurf_vis}, we show the visualization of a representative isosurface generated from Isabel dataset using the Velocity field with isovalue=41. We observe that the MVNet and TTHRESH produce visually comparable results, while the isosurface generated by the LERP fails to capture detailed surface features.
\begin{table}[!thb]
\centering
\caption{Quantitative feature quality comparison using average Chamfer and Hausdorff distances for all data variables for two 3D datasets. The findings indicate that for both the datasets, MVNet outperforms TTHRESH and LERP regarding isocontour reconstruction quality.}
\label{cf_table}
\resizebox{\linewidth}{!}{
\begin{tabular}{|c|cc|cc|cc|}
\hline
\textbf{Dataset} & \multicolumn{2}{c|}{\textbf{MVNet}} & \multicolumn{2}{c|}{\textbf{TTHRESH}} & \multicolumn{2}{c|}{\textbf{LERP}} \\ \hline
                     & \multicolumn{1}{c|}{\textbf{Chamfer $\downarrow$}}  & \textbf{Hausdorff $\downarrow$} & \multicolumn{1}{c|}{\textbf{Chamfer $\downarrow$}}  & \textbf{Hausdorff $\downarrow$} & \multicolumn{1}{c|}{\textbf{Chamfer $\downarrow$}}   & \textbf{Hausdorff $\downarrow$} \\ \hline
\textbf{Combustion }& \multicolumn{1}{c|}{0.250}  & 11.937 & \multicolumn{1}{c|}{0.344} & 13.813 & \multicolumn{1}{c|}{5.715}  & 32.187 \\ \hline
\textbf{Isabel  }   & \multicolumn{1}{c|}{0.955} & 23.016 & \multicolumn{1}{c|}{2.814} & 36.131 & \multicolumn{1}{c|}{12.959} & 60.941 \\ \hline
\end{tabular}
}
\end{table}

\begin{table}[thb]
\caption{Visualization quality comparison in image space. We use SSIM, LPIPS, and DISTS measures to quantify perceptual image similarity between ground truth and the images generated by MVNet, TTHRESH, and LERP methods.}
\label{vis_compare_table}
\resizebox{\linewidth}{!}{%
\begin{tabular}{|c|c|ccc|ccc|ccc|}
\hline
\multirow{2}{*}{\textbf{Dataset}} &
  \multirow{2}{*}{\textbf{Var Name}} &
  \multicolumn{3}{c|}{\textbf{SSIM $\uparrow$}} &
  \multicolumn{3}{c|}{\textbf{LPIPS $\downarrow$}} &
  \multicolumn{3}{c|}{\textbf{DISTS $\downarrow$}} \\ \cline{3-11} 
 &
   &
  \multicolumn{1}{c|}{\textbf{MVNet}} &
  \multicolumn{1}{c|}{\textbf{TTHRESH}} &
  \textbf{LERP} &
  \multicolumn{1}{c|}{\textbf{MVNet}} &
  \multicolumn{1}{c|}{\textbf{TTHRESH}} &
  \textbf{LERP} &
  \multicolumn{1}{c|}{\textbf{MVNet}} &
  \multicolumn{1}{c|}{\textbf{TTHRESH}} &
  \textbf{LERP} \\ \hline
\multirow{2}{*}{\textbf{Isabel}} &
  P &
  \multicolumn{1}{c|}{0.991} &
  \multicolumn{1}{c|}{0.936} &
  0.675 &
  \multicolumn{1}{c|}{0.049} &
  \multicolumn{1}{c|}{0.248} &
  0.356 &
  \multicolumn{1}{c|}{0.053} &
  \multicolumn{1}{c|}{0.224} &
  0.243 \\ \cline{2-11} 
 &
  Vel &
  \multicolumn{1}{c|}{0.956} &
  \multicolumn{1}{c|}{0.95} &
  0.858 &
  \multicolumn{1}{c|}{0.085} &
  \multicolumn{1}{c|}{0.075} &
  0.174 &
  \multicolumn{1}{c|}{0.081} &
  \multicolumn{1}{c|}{0.115} &
  0.299 \\ \hline
\multirow{2}{*}{\textbf{Climate50}} &
  U10 &
  \multicolumn{1}{c|}{0.975} &
  \multicolumn{1}{c|}{0.914} &
  0.744 &
  \multicolumn{1}{c|}{0.003} &
  \multicolumn{1}{c|}{0.023} &
  0.155 &
  \multicolumn{1}{c|}{0.028} &
  \multicolumn{1}{c|}{0.221} &
  0.15 \\ \cline{2-11} 
 &
  CLDTOT &
  \multicolumn{1}{c|}{0.987} &
  \multicolumn{1}{c|}{0.967} &
  0.88 &
  \multicolumn{1}{c|}{0.005} &
  \multicolumn{1}{c|}{0.014} &
  0.137 &
  \multicolumn{1}{c|}{0.036} &
  \multicolumn{1}{c|}{0.191} &
  0.198 \\ \hline
\end{tabular}%
}
\end{table}

\begin{figure}[!tb]
\centering
\begin{subfigure}[t]{0.23\linewidth}
    \centering
    \includegraphics[width=\linewidth]{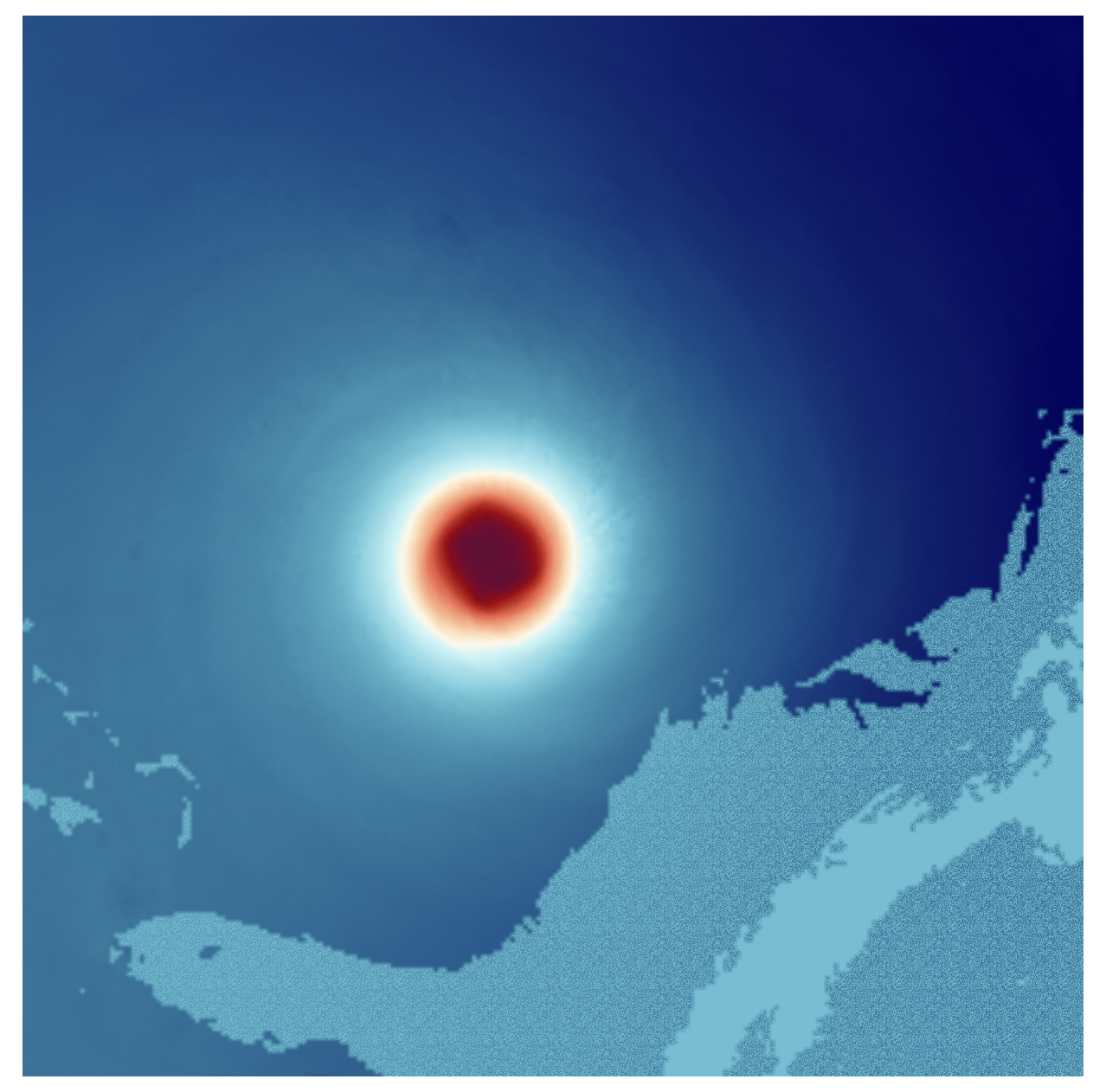}
    \caption{GT}
    \label{Isabel_P_GT}
\end{subfigure}
~
\begin{subfigure}[t]{0.23\linewidth}
    \centering
    \includegraphics[width=\linewidth]{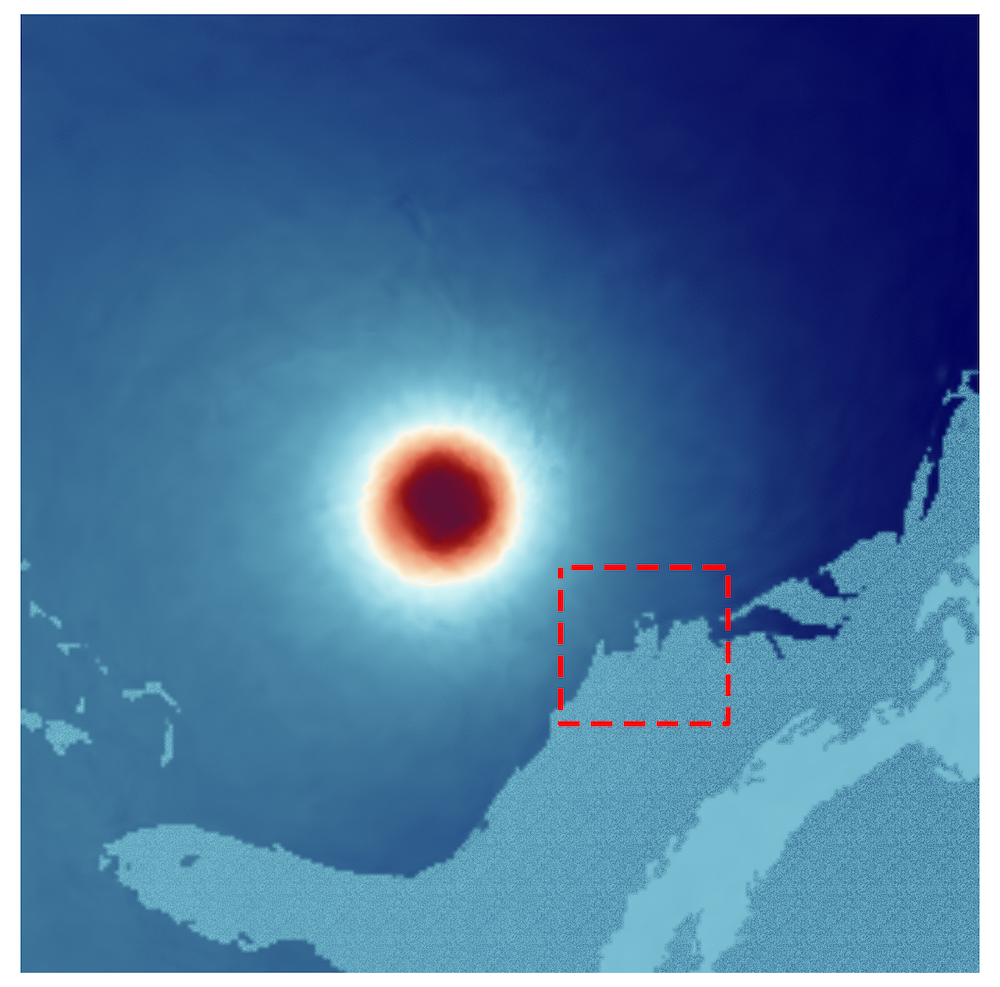}
    \caption{MVNet}
    \label{Isabel_P_MVNet}
\end{subfigure}
~
\begin{subfigure}[t]{0.23\linewidth}
    \centering
    \includegraphics[width=\linewidth]{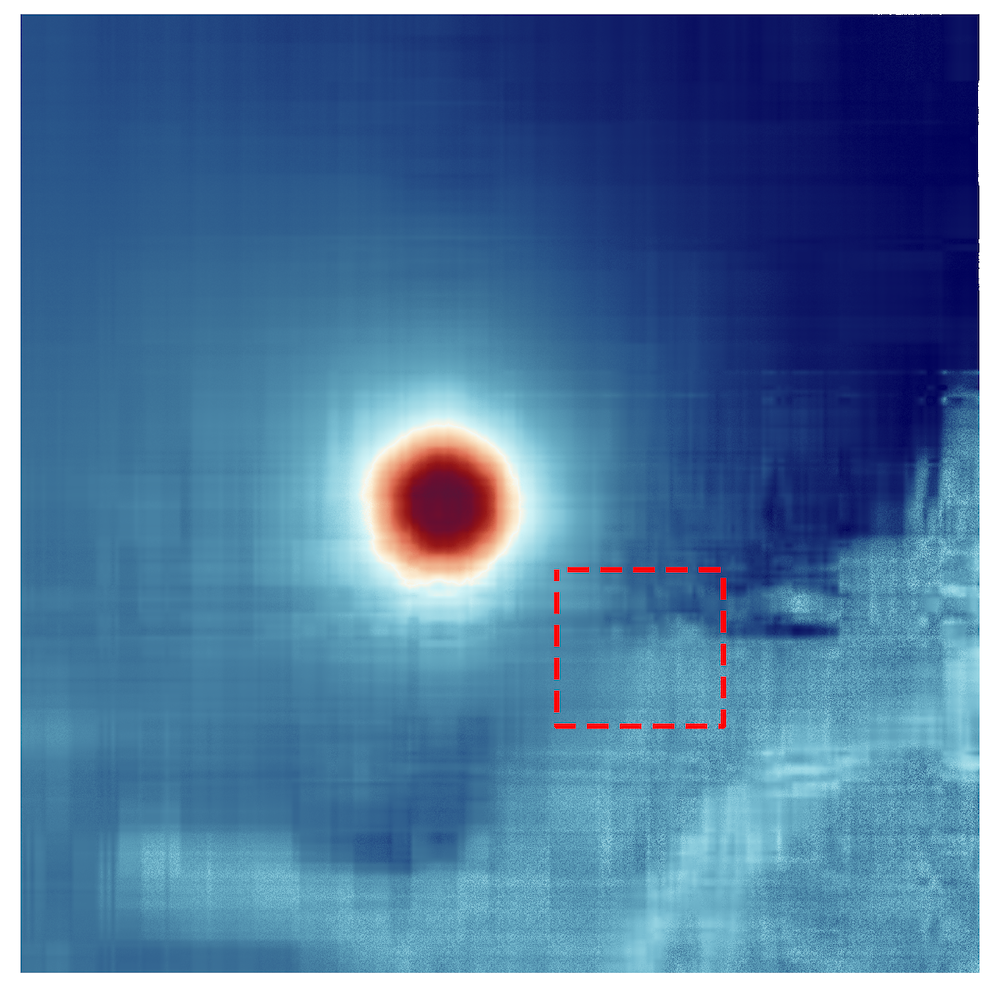}
    \caption{TTHRESH}
    \label{Isabel_P_TTHRESH}
\end{subfigure}
~
\begin{subfigure}[t]{0.23\linewidth}
    \centering
    \includegraphics[width=\linewidth]{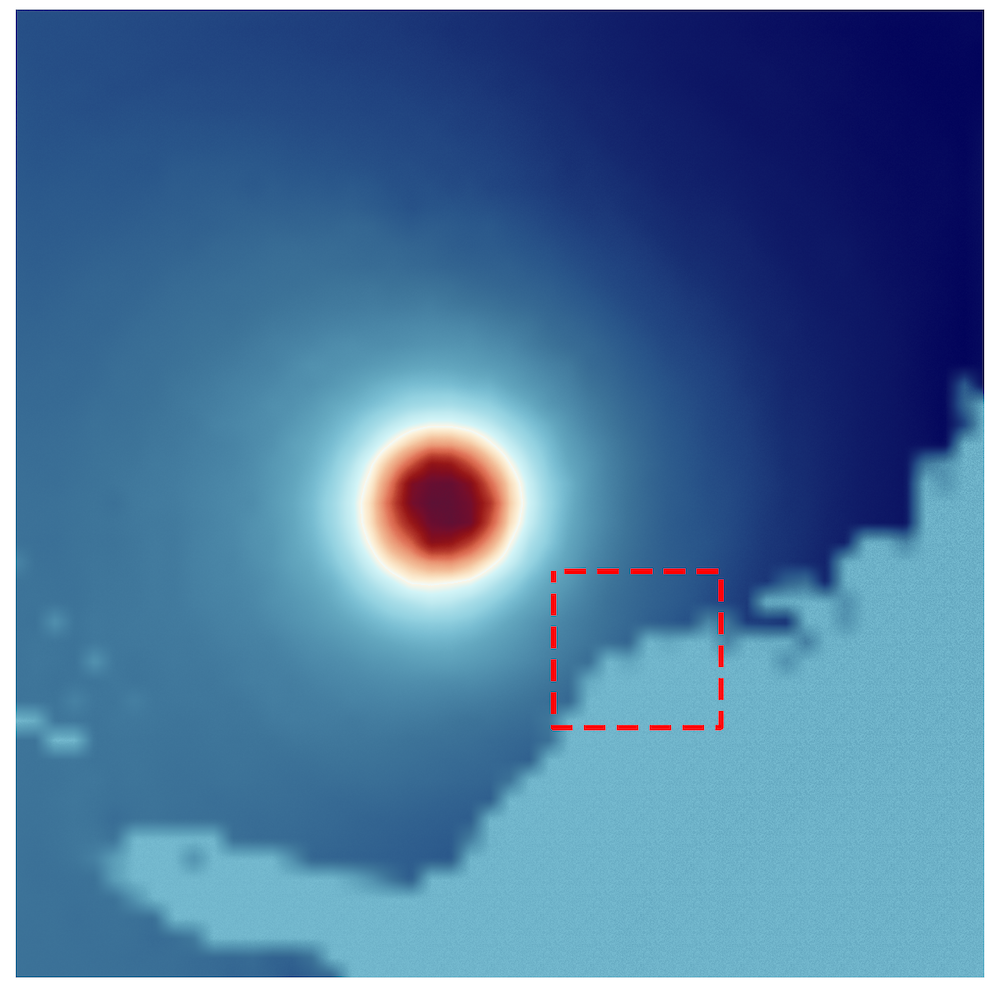}
    \caption{LERP}
    \label{Isabel_P_LERP}
\end{subfigure}
\caption{Reconstructed data visualization for Pressure (P) variable of Isabel dataset. Images generated by ground truth (GT), MVNet, TTHRESH, and LERP are shown. We observe that MVNet produces the most accurate result.}
\label{isabel_vis_comp}
\end{figure}

\begin{figure*}[thb]
\centering
\begin{subfigure}[t]{0.22\linewidth}
    \centering
    \includegraphics[width=\linewidth]{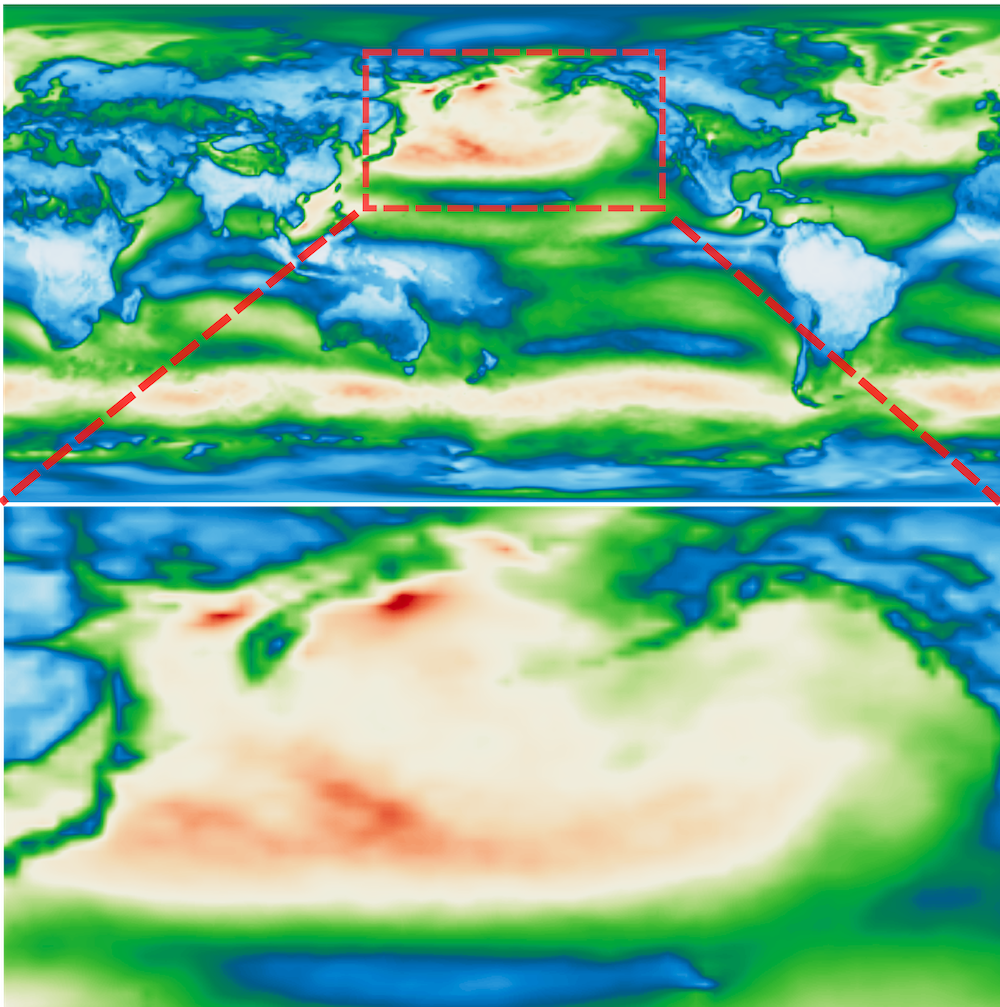}
    \caption{Ground truth}
    \label{climate100_u10_GT_comb}
\end{subfigure}
~
\begin{subfigure}[t]{0.22\linewidth}
    \centering
    \includegraphics[width=\linewidth]{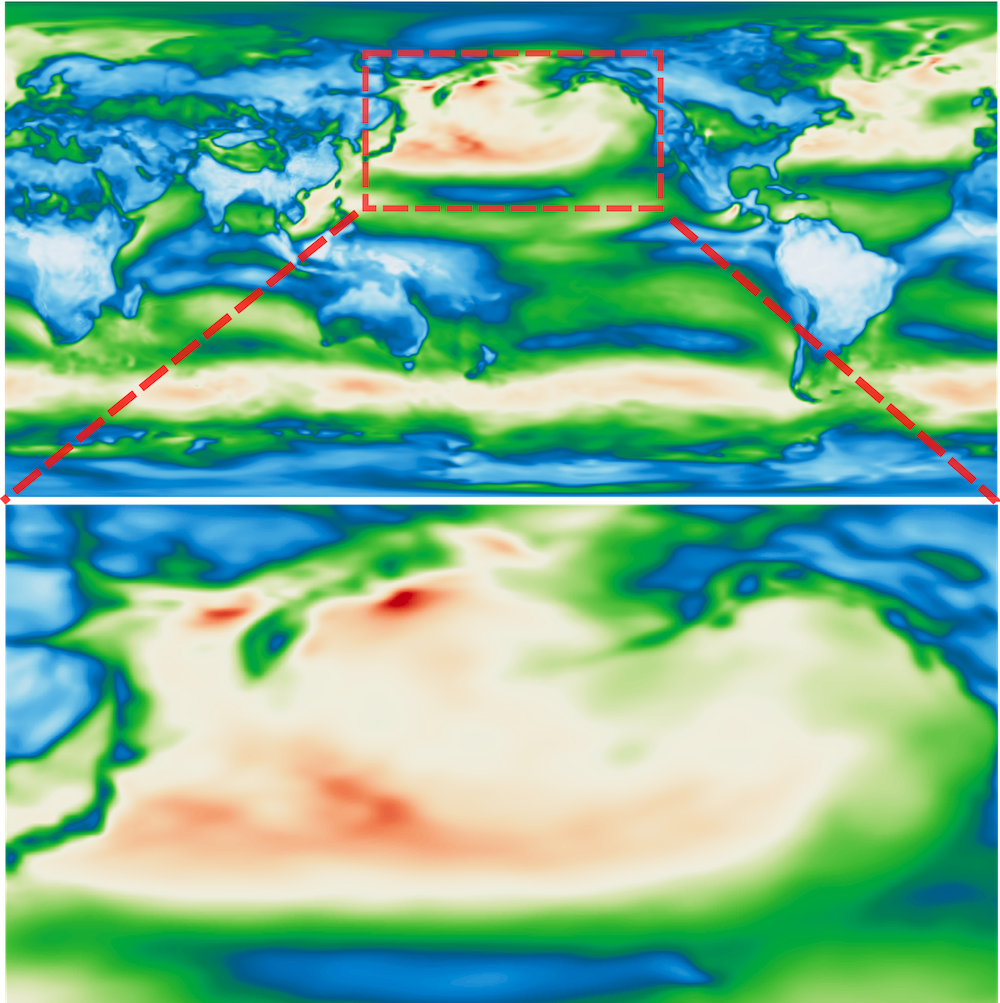}
    \caption{MVNet}
    \label{climate100_u10_MVNet_comb}
\end{subfigure}
~
\begin{subfigure}[t]{0.22\linewidth}
    \centering
    \includegraphics[width=\linewidth]{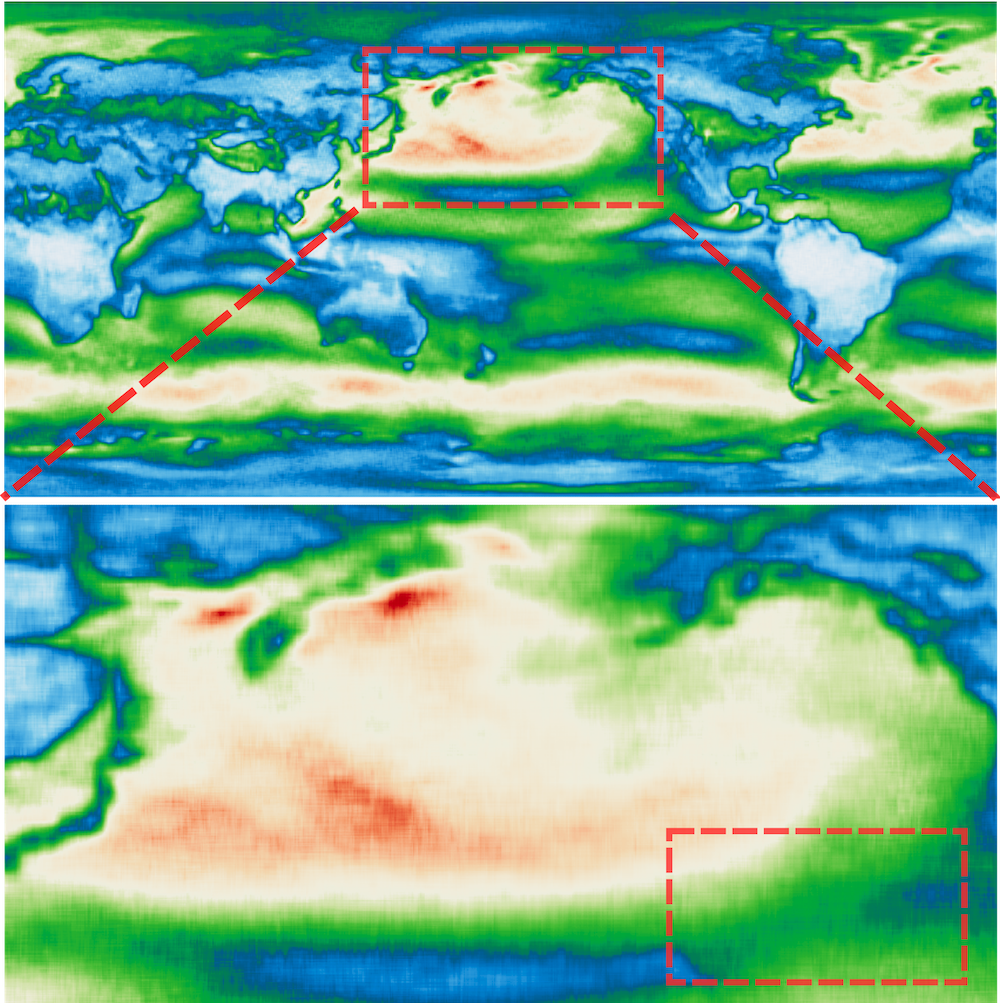}
    \caption{TTHRESH}
    \label{climate100_u10_TTHRESH_comb}
\end{subfigure}
~
\begin{subfigure}[t]{0.22\linewidth}
    \centering
    \includegraphics[width=\linewidth]{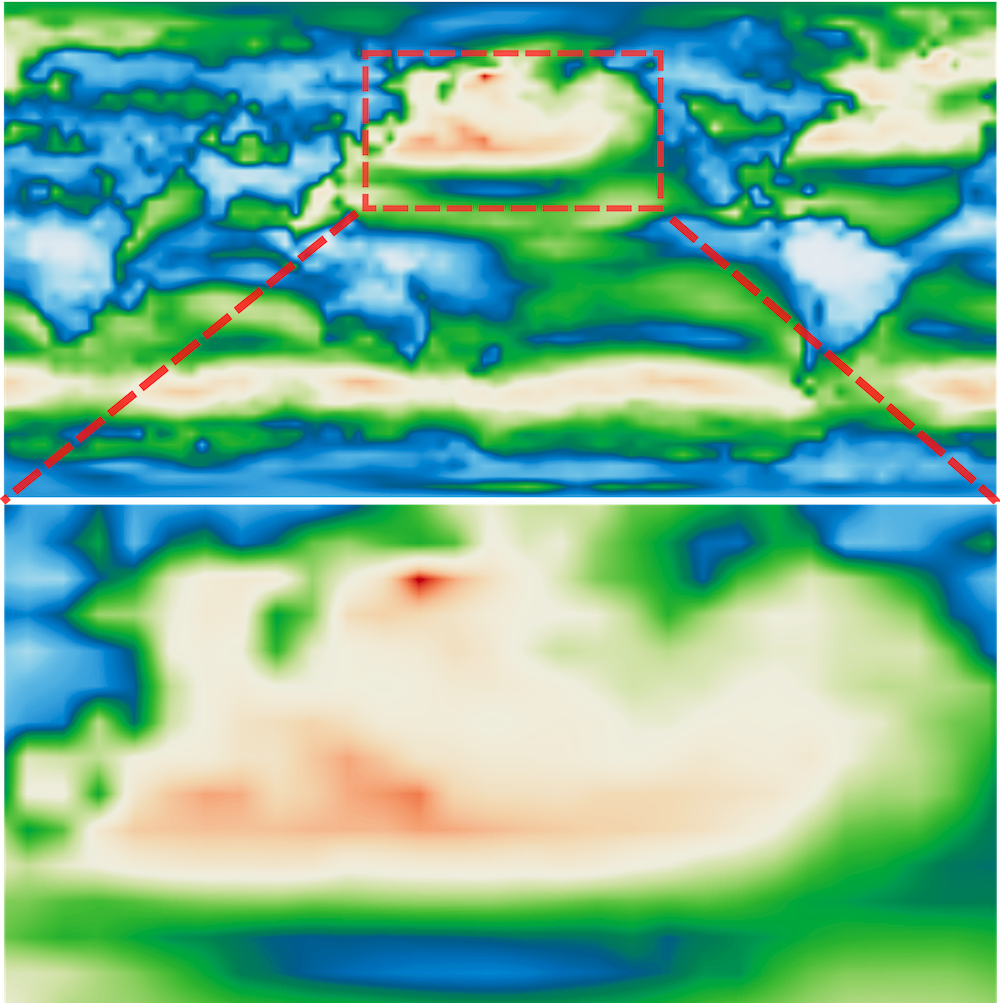}
    \caption{LERP}
    \label{climate100_u10_LERP_comb}
\end{subfigure}
\caption{Reconstructed data visualization for the U-Velocity (U10) variable of the Climate100 dataset. Images generated by ground truth data, MVNet, TTHRESH, and LERP methods are shown from left to right with a region zoomed below to show the differences. We observe that MVNet produces the most accurate result, while the result from TTHRESH shows artifacts as highlighted by the red dotted box in Fig.~\ref{climate100_u10_TTHRESH_comb}. The LERP method-generated image is the least accurate.}
\label{climate100_vis_comp}
\end{figure*}

\subsection{Comparison of Visualization Quality}
Next comparative evaluation is conducted in image space to study the accuracy of the visualizations generated. We use selected variables from Isabel and Climate50 data to conduct this study. While generating images, we keep all the rendering settings fixed for each dataset. To quantify the image similarity between the ground truth and images generated by each method, we use (1) Structural  Similarity (SSIM)~\cite{ssim}, (2) Learned Perceptual Image Patch Similarity (LPIPS)~\cite{lpips}, and (3) Deep Image Structure And Texture Similarity (DISTS)~\cite{dists}. The results are presented in Table~\ref{vis_compare_table}. While MVNet produces the most accurate visualizations, the LERP consistently yields the least accurate visualizations. For a qualitative comparison, in Fig~\ref{isabel_vis_comp} and Fig~\ref{climate100_vis_comp}, we provide visualizations of selected variables from Isabel and Climate100 dataset respectively. From Fig~\ref{isabel_vis_comp}, we observe that MVNet produces the most accurate result while TTHRESH and LERP-generated images contain visual artifacts. Fig~\ref{climate100_vis_comp} shows a similar finding as Fig~\ref{isabel_vis_comp}. By observing the zoomed inset, we see that MVNet produces the most accurate result for the U10 variable and TTHRESH-generated image contains visual artifacts. The LERP method fails to preserve the features and produces the least accurate result.

\subsection{Comparison with Copula-based Method}
\begin{figure}[htb]
\centering
\includegraphics[width=0.6\linewidth]{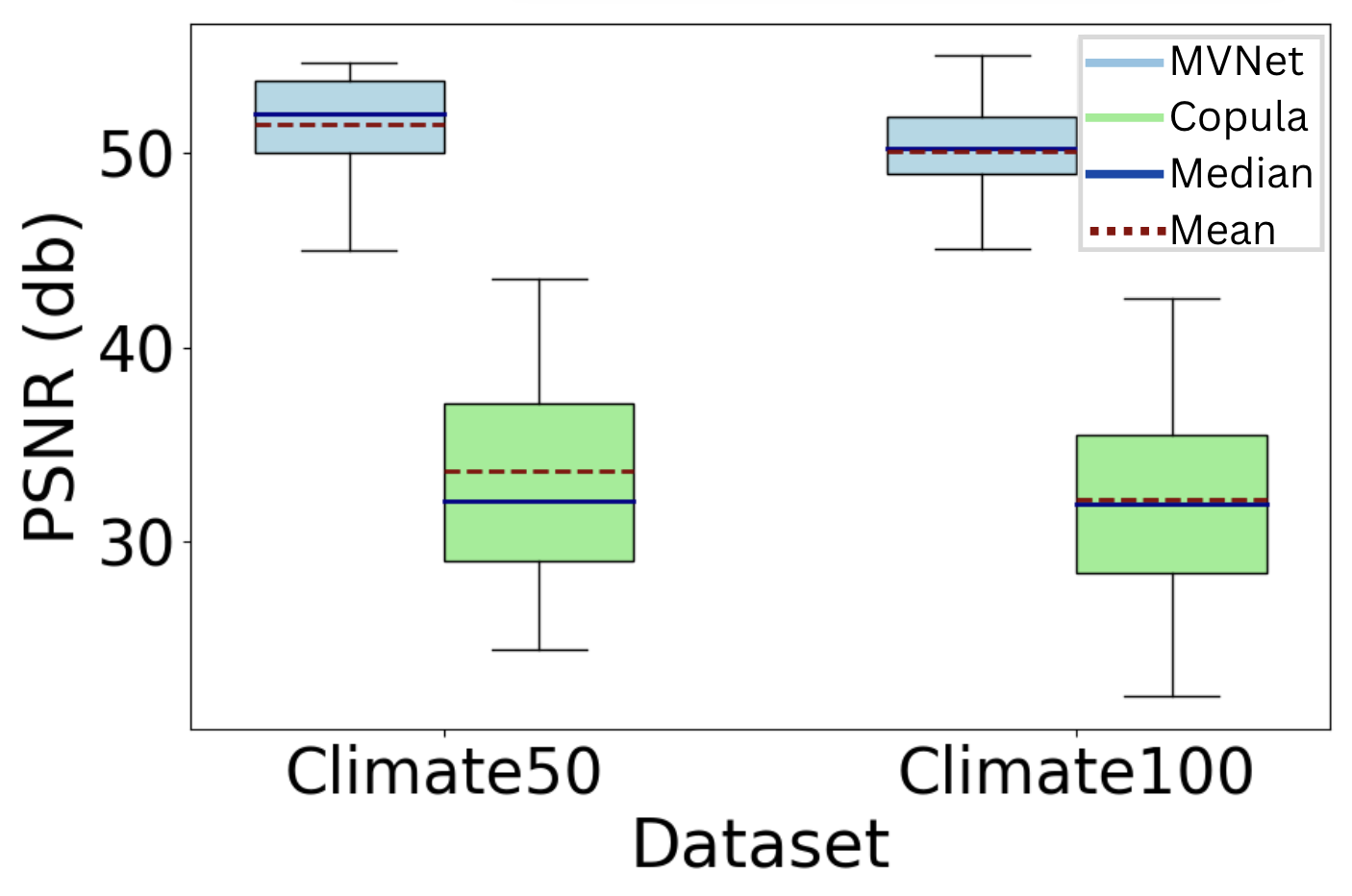}
\caption{Boxplot-based comparison of PSNR for MVNet and Copula method for Climate50 and Climate100 datasets.}
\label{boxplot_copula}
\end{figure}

Representing multivariate datasets with a large number of variables compactly using statistical Copula-based summaries was proposed by Hazarika et al.~\cite{codda}. Since our method is addressing the similar problem, we use the Climate50 and Climate100 datasets to compare MVNet with the Copula-based summarization method~\cite{codda}. To keep the storage footprint minimal for the Copula-based model, we model each variable in a block using Gaussian distribution, which only requires two floating points per variable per block. We also store the pairwise correlation coefficient for all pairs of variables for each data block. To make the storage size comparable to MVNet, we use $80\times80$ block size for the Climate50 dataset and $120\times120$ block size for the Climate100 dataset for Copula modeling. For both MVNet and Copula-based methods, we reconstruct all the variables at their original resolution. In Table~\ref{copula_table}, the quality (average PSNR across all variables) vs. storage trade-off is presented. We observe that MVNet produces significantly higher-quality reconstruction with lower storage (as shown in boxplot of Fig.~\ref{boxplot_copula}). Due to space limitation, we show visualization of a representative variable, Surface Temperature (TS), in the supplementary material.

\begin{table}[tbh]
\centering
\caption{Quantitative comparison of the data reconstruction quality between MVNet and Copula-based method using Climate dataset. It is observed that MVNet produces higher reconstruction quality with lower storage overhead.}
\label{copula_table}
\resizebox{\linewidth}{!}{
\begin{tabular}{|cc|cc|cccl|}
\hline
\multicolumn{2}{|c|}{}                 & \multicolumn{2}{c|}{\textbf{MVNet}}            & \multicolumn{4}{c|}{\textbf{Copula-based Method}}                                                      \\ \hline
\multicolumn{1}{|c|}{\textbf{dataset}} &
  \textbf{\#Variables} &
  \multicolumn{1}{c|}{\begin{tabular}[c]{@{}c@{}}\textbf{Storage} $\downarrow$\\ (KB)\end{tabular}} &
  \begin{tabular}[c]{@{}c@{}}\textbf{PSNR} $\uparrow$\\ (dB)\end{tabular} &
  \multicolumn{1}{c|}{\begin{tabular}[c]{@{}c@{}}\textbf{Block}\\ \textbf{Size}\end{tabular}} &
  \multicolumn{1}{c|}{\begin{tabular}[c]{@{}c@{}}\textbf{Storage} $\downarrow$\\ (KB)\end{tabular}} &
  \multicolumn{2}{c|}{\begin{tabular}[c]{@{}c@{}}\textbf{PSNR} $\uparrow$\\ (dB)\end{tabular}} \\ \hline
\multicolumn{1}{|c|}{Climate50}  & 53  & \multicolumn{1}{c|}{1184} & 51.274 & \multicolumn{1}{c|}{80X80}   & \multicolumn{1}{c|}{5116} & \multicolumn{2}{c|}{33.618} \\ \hline
\multicolumn{1}{|c|}{Climate100} & 100 & \multicolumn{1}{c|}{1208} & 49.679 & \multicolumn{1}{c|}{120X120} & \multicolumn{1}{c|}{7108} & \multicolumn{2}{c|}{32.192} \\ \hline
\end{tabular}
}
\end{table}

\section{Multivariate Application Study}
\subsection{Multivariate Dependency Analysis}
Variables in multivariate scientific datasets are often correlated with other variables in the dataset. The nature of this correlation or interdependence can be both linear and non-linear~\cite{vis13_info_theory}. By analyzing and visualizing such correlated (dependent) variables, complex multivariate features are often studied~\cite{codda, corr_vis, corr_vis_1, vis13_info_theory}. Hence, it is essential to preserve such correlations for any multivariate data modeling technique. In this section, we compare the accuracy with which correlation (linear and non-linear) is preserved in datasets reconstructed by the MVNet, TTHRESH, and LERP methods. We use Pearson's Correlation Coefficient to quantify linear dependence and Mutual Information to measure non-linear dependence between all variable pairs for each method. Next, to compute the error in these estimated correlation and mutual information values, we compute the absolute difference between each estimated value and the ground truth value for each variable pair. In Table~\ref{corr_mi_table}, we show the average absolute error value computed over the three methods' variable pairs for each dataset. 

\begin{figure*}[thb]
\centering
\begin{subfigure}[t]{0.15\linewidth}
    \centering
    \includegraphics[width=\linewidth]{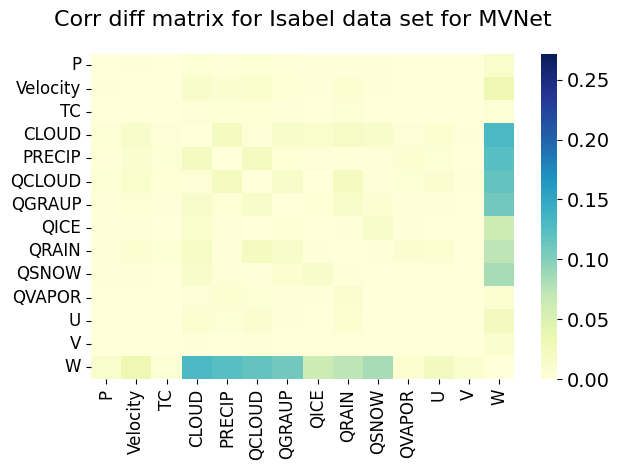}
    \caption{\raggedright Corr diff matrix of Isabel for MVNet}
    \label{}
\end{subfigure}
~
\begin{subfigure}[t]{0.15\linewidth}
    \centering
    \includegraphics[width=\linewidth]{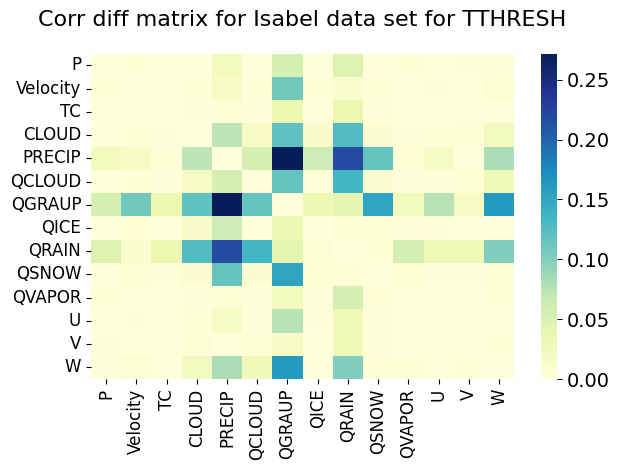}
    \caption{\raggedright Corr diff matrix of Isabel for TTHRESH}
    \label{}
\end{subfigure}
~
\begin{subfigure}[t]{0.15\linewidth}
    \centering
    \includegraphics[width=\linewidth]{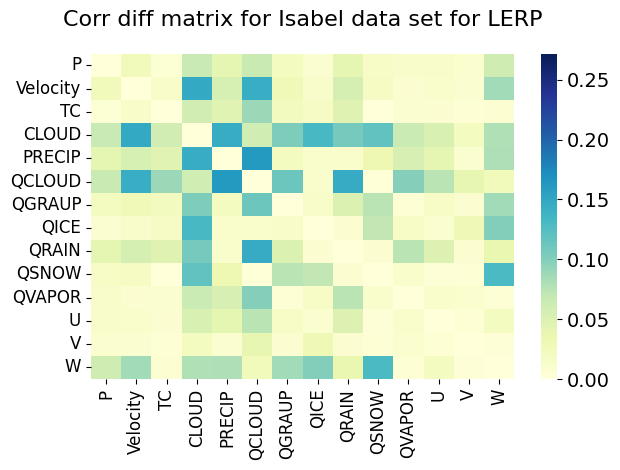}
    \caption{\raggedright Corr diff matrix for Isabel for LERP}
    \label{}
\end{subfigure}
~
\begin{subfigure}[t]{0.15\linewidth}
    \centering
    \includegraphics[width=\linewidth]{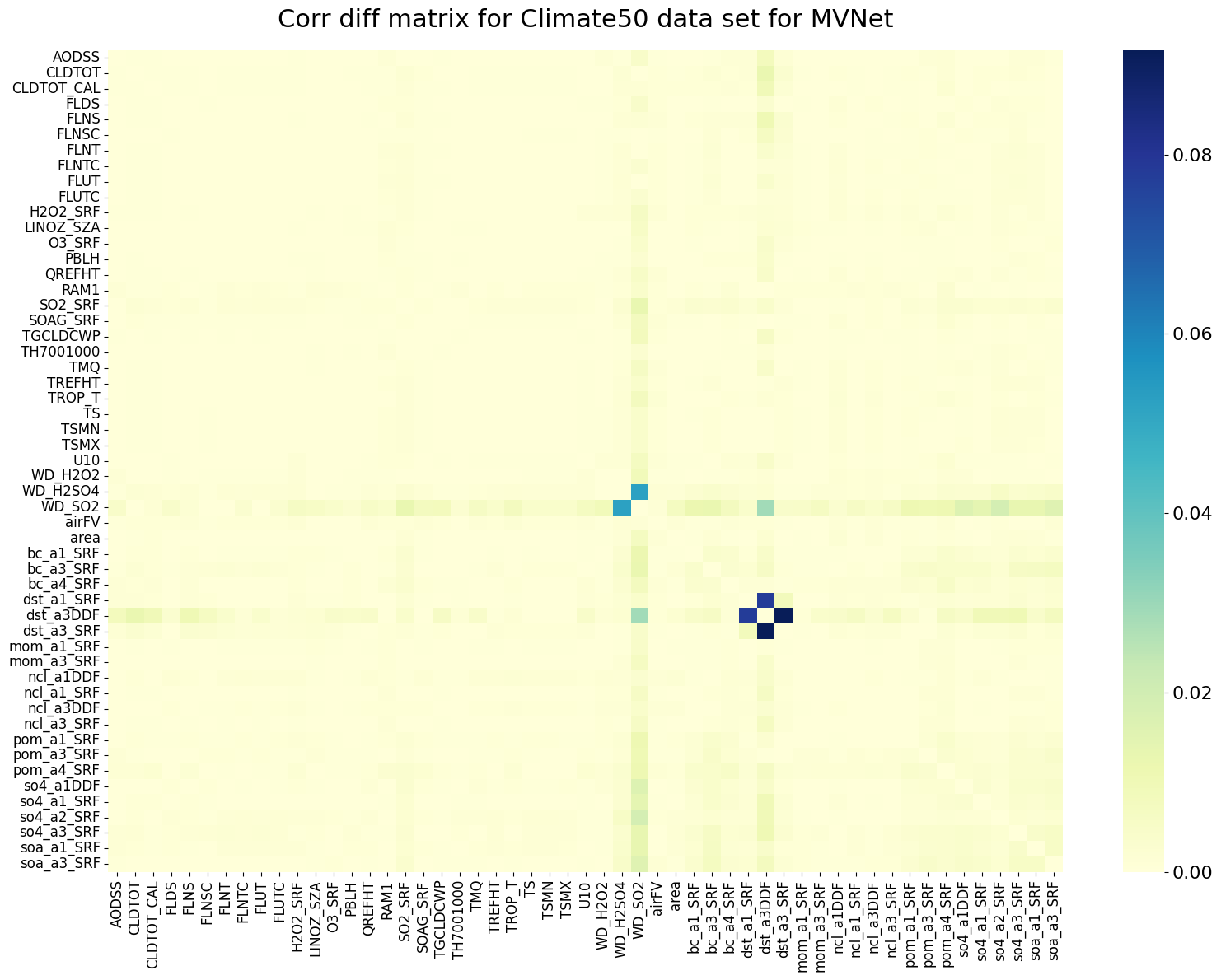}
    \caption{\raggedright Corr diff matrix for clim50 for MVNet}
    \label{}
\end{subfigure}
~
\begin{subfigure}[t]{0.15\linewidth}
    \centering
    \includegraphics[width=\linewidth]{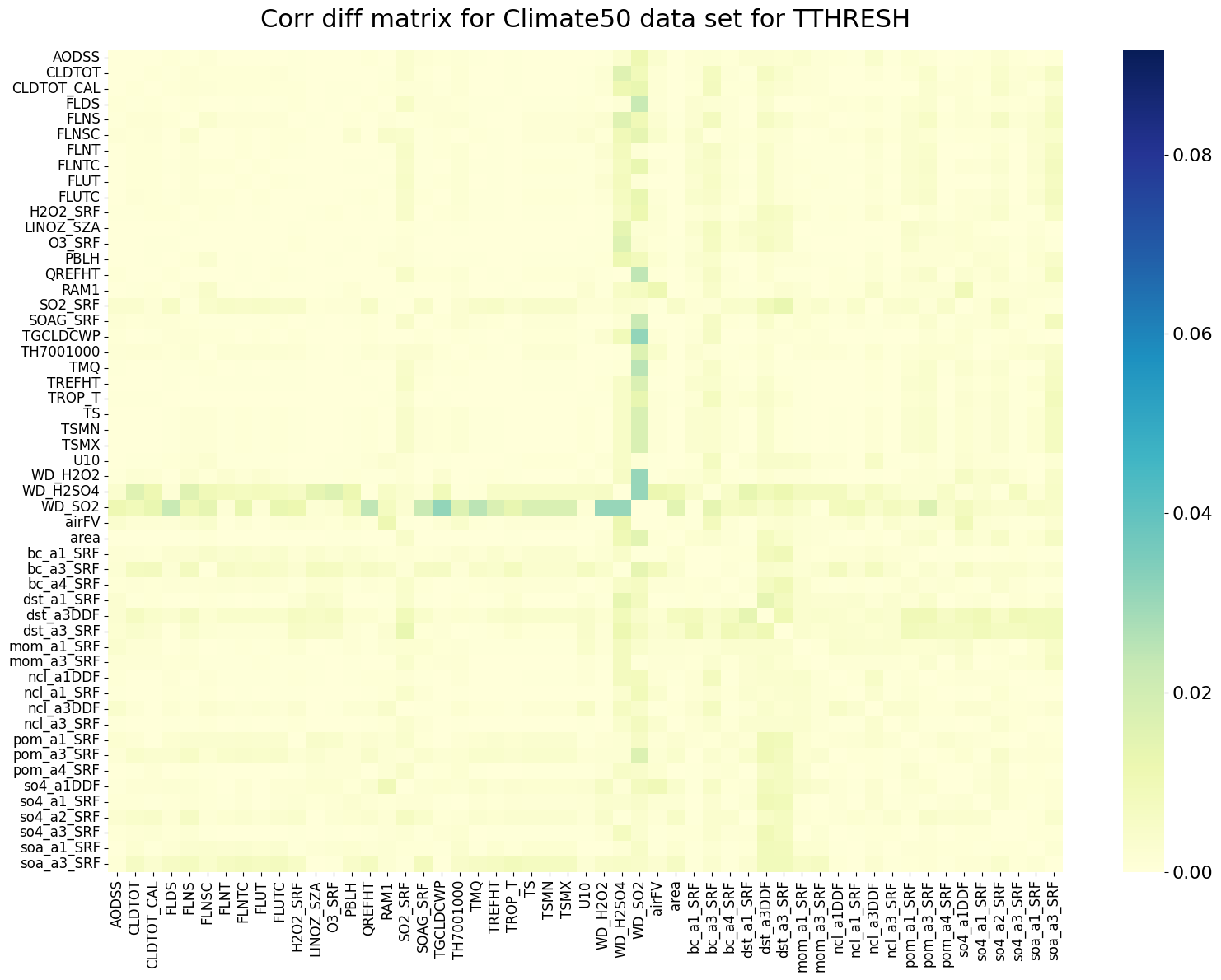}
    \caption{\raggedright Corr diff matrix of clim50 for TTHRESH}
    \label{}
\end{subfigure}
~
\begin{subfigure}[t]{0.15\linewidth}
    \centering
    \includegraphics[width=\linewidth]{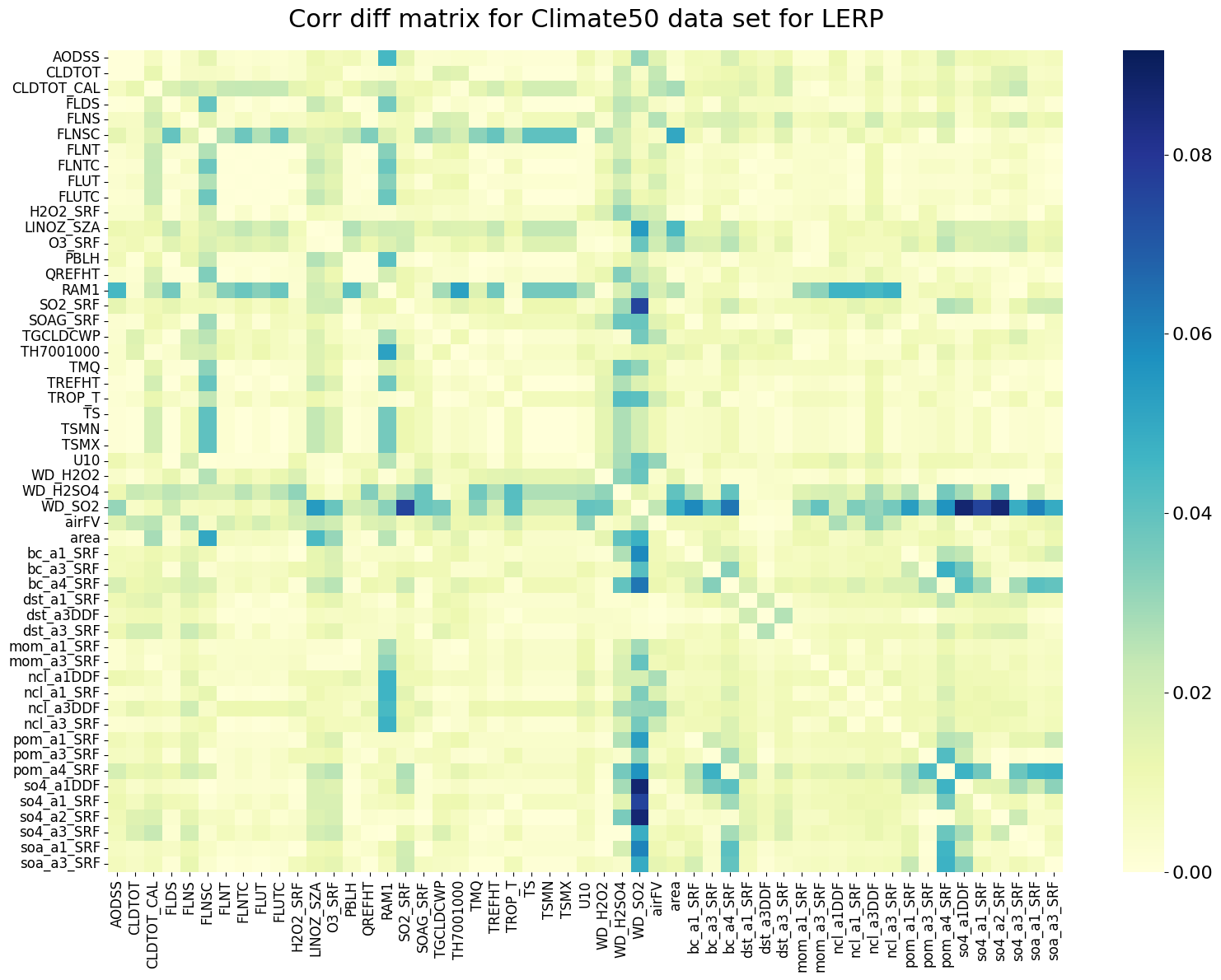}
    \caption{\raggedright Corr diff matrix for clim50 for LERP}
    \label{}
\end{subfigure}
\caption{Linear Correlation Error Matrices for Isabel and Climate50 dataset for MVNet, TTHRESH, and LERP method.}
\label{corr_isabel_turbine}
\end{figure*}

\begin{figure*}[thb]
    \centering
    \begin{subfigure}[t]{0.15\linewidth}
        \centering
        \includegraphics[width=\linewidth]{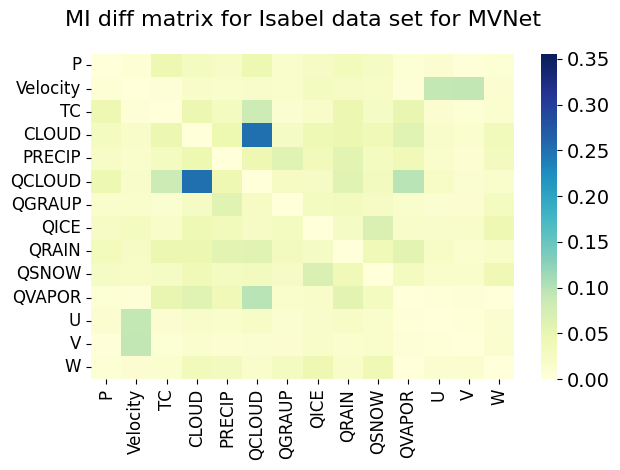}
        \caption{\raggedright MI diff matrix for Isabel for MVNet}
        \label{}
    \end{subfigure}
    ~
    \begin{subfigure}[t]{0.15\linewidth}
        \centering
        \includegraphics[width=\linewidth]{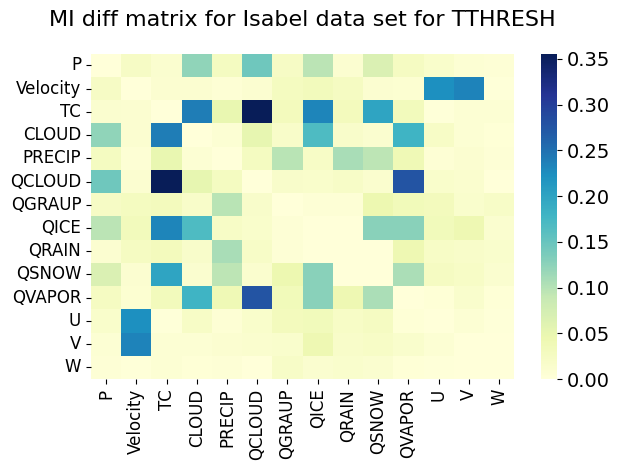}
        \caption{\raggedright MI diff matrix for Isabel for TTHRESH}
        \label{}
    \end{subfigure}
    ~
    \begin{subfigure}[t]{0.15\linewidth}
        \centering
        \includegraphics[width=\linewidth]{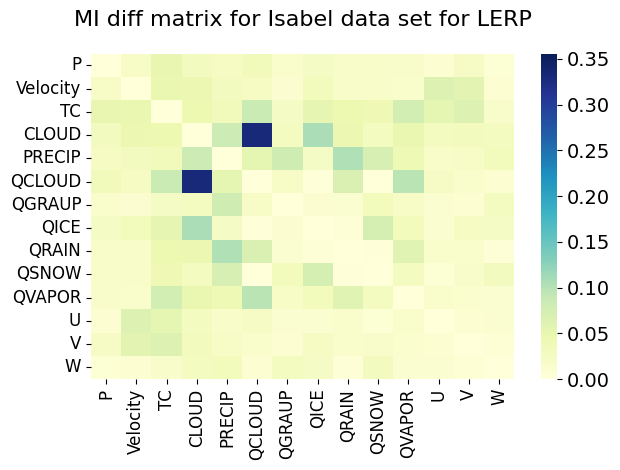}
        \caption{\raggedright MI diff matrix for Isabel for LERP}
        \label{}
    \end{subfigure}
    ~
    \begin{subfigure}[t]{0.15\linewidth}
        \centering
        \includegraphics[width=\linewidth]{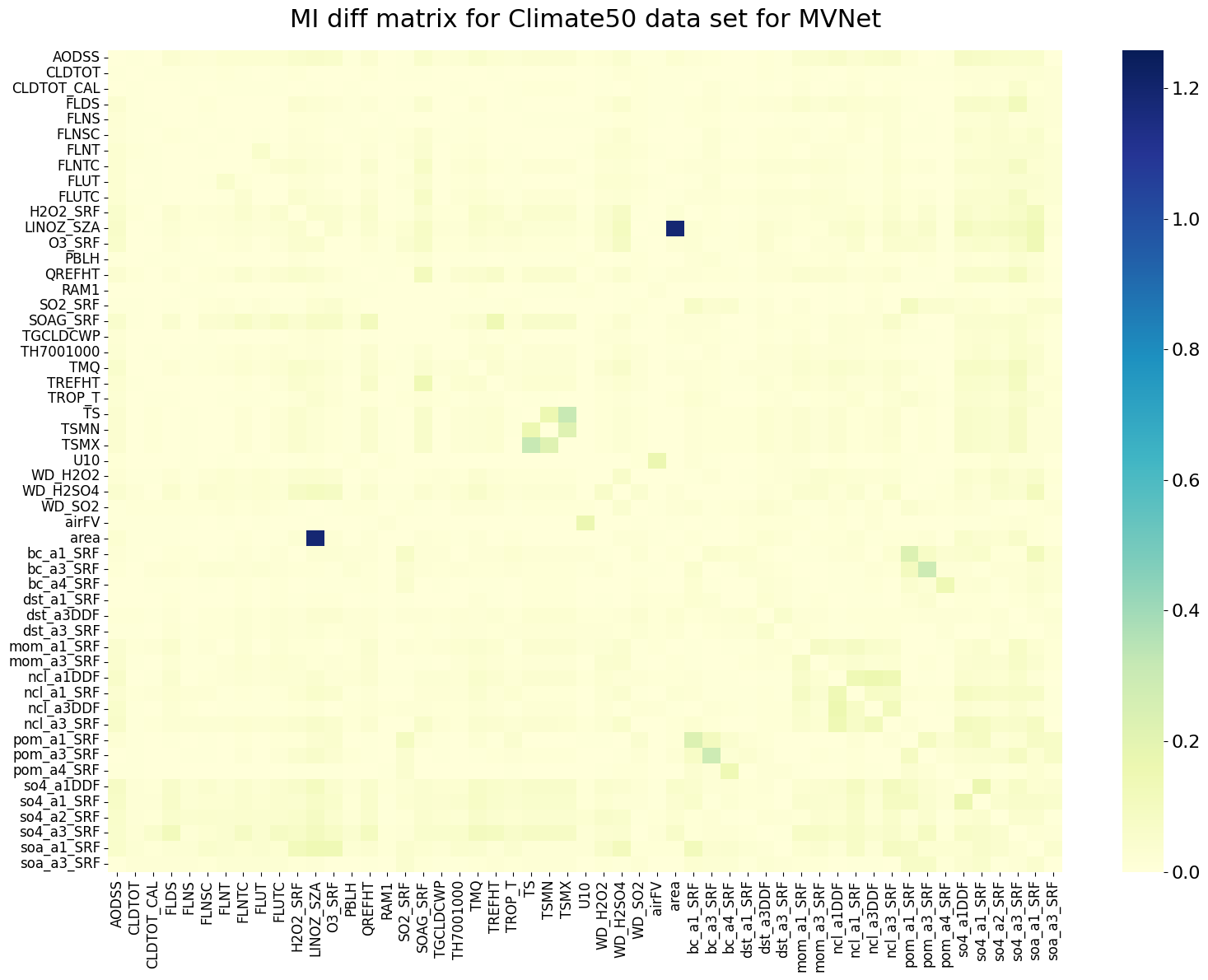}
        \caption{\raggedright MI diff matrix for climate50  for MVNet}
        \label{}
    \end{subfigure}
    ~
    \begin{subfigure}[t]{0.15\linewidth}
        \centering
        \includegraphics[width=\linewidth]{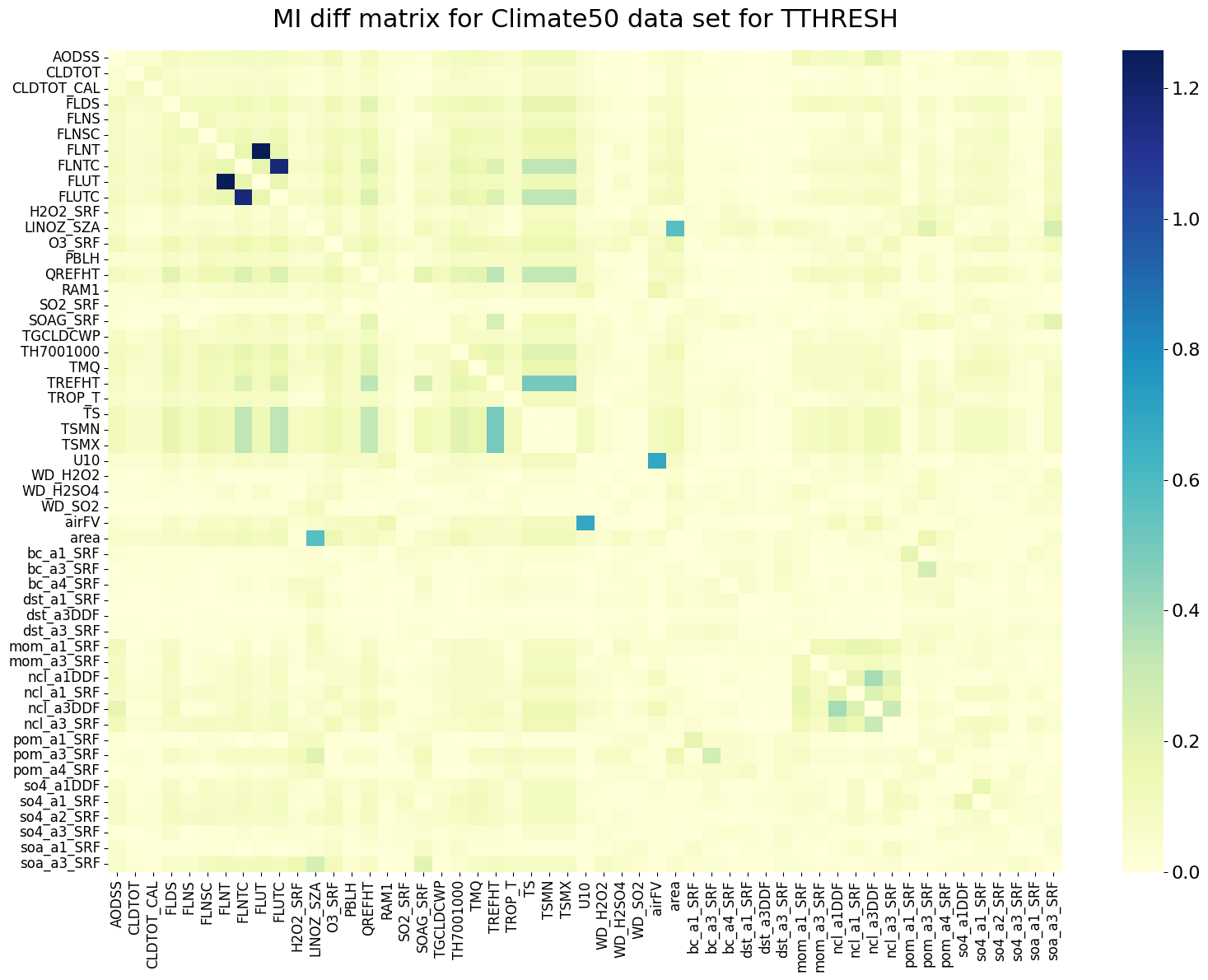}
        \caption{\raggedright MI diff matrix for clim50  for TTHRESH}
        \label{}
    \end{subfigure}
    ~
    \begin{subfigure}[t]{0.15\linewidth}
        \centering
        \includegraphics[width=\linewidth]{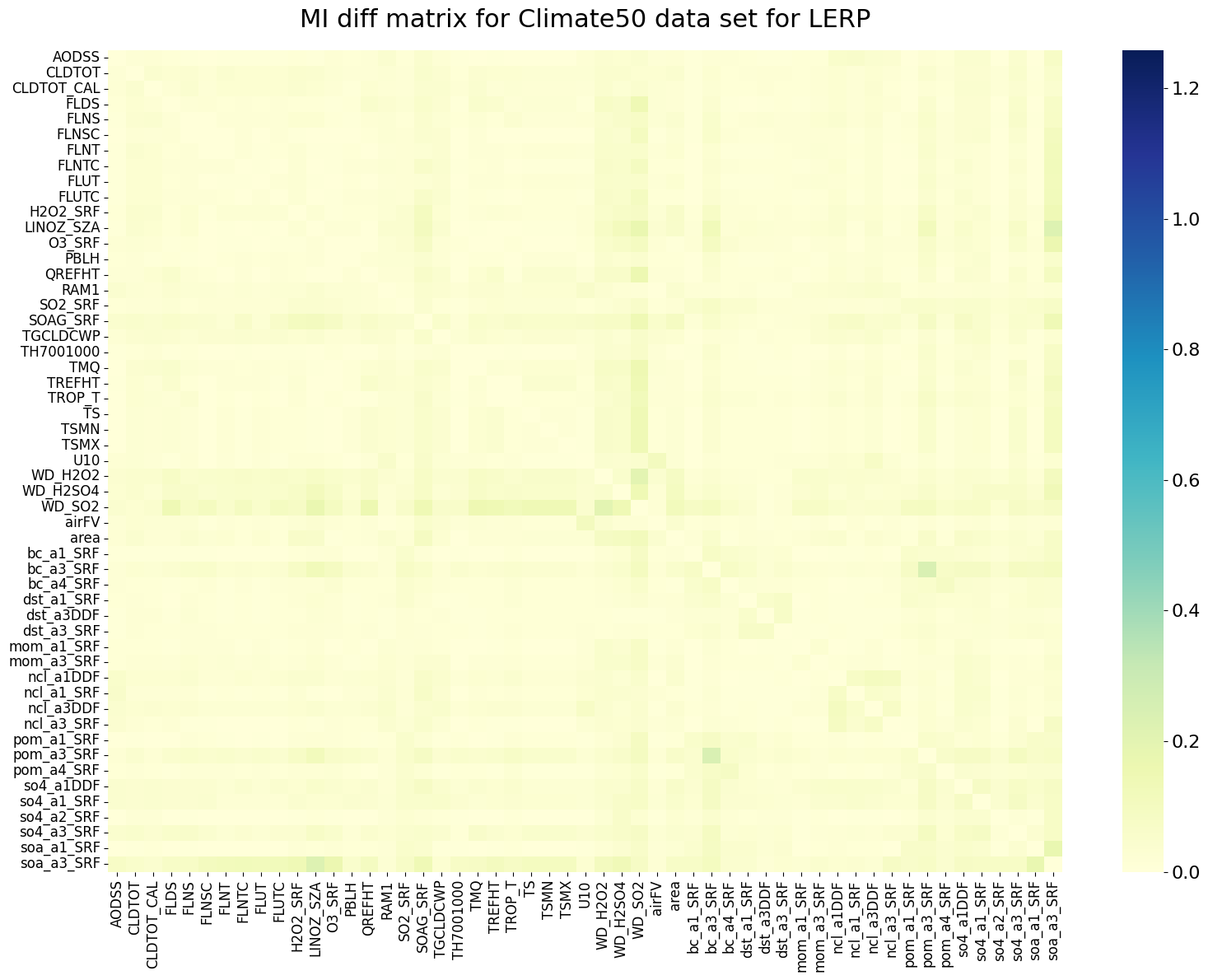}
        \caption{\raggedright MI diff matrix for climate50  for LERP}
        \label{}
    \end{subfigure}
    \caption{\raggedright Mutual Information Error Matrices for Isabel and Climate50 dataset for MVNet, TTHRESH, and LERP method.}
    \label{mi_isabel_turbine}
\end{figure*}

It is observed that, in terms of linear correlation error (deviation), for Combustion data, TTHRESH incurs the least average error, and for all other datasets, MVNet produces the minimum average error. Next, it is found that MVNet produces the least average absolute error in mutual information for all the datasets. LERP produces the highest error. Therefore, from Table~\ref{corr_mi_table}, we conclude that MVNet is recovers the dependencies among variables most accurately. In Fig.~\ref{corr_isabel_turbine} and Fig.~\ref{mi_isabel_turbine}, we show all pairwise correlation and mutual information error matrices for Isabel and Climate50 datasets, respectively. Since we show the absolute error values in these matrices, if a matrix has fewer darker cells, then it can be interpreted that the corresponding method produces fewer errors. We use a consistent color map while generating these plots across the three methods so that the same color intensity in two different plots indicates the same absolute error value. We observe that between Figs.~\ref{corr_isabel_turbine}(a)-(c), Fig.~\ref{corr_isabel_turbine}(a), and between Figs.~\ref{corr_isabel_turbine}(d)-(f), Fig.~\ref{corr_isabel_turbine}(d) produces a minimum error indicating MVNet's superiority over TTHRESH and LERP in estimating linear correlation. Similarly, from Fig.~\ref{mi_isabel_turbine}, by comparing Figs.~\ref{mi_isabel_turbine}(a)-(c) and Figs.~\ref{mi_isabel_turbine}(d)-(f), we conclude that MVNet produces the most accurate mutual information estimation as compared to TTHRESH and LERP.

\begin{table}[!t]
\centering
\caption{Comparison of average absolute error in linear correlation and mutual information for different methods.}
\label{corr_mi_table}
\resizebox{\linewidth}{!}{
\begin{tabular}{|c|ccc|ccc|}
\hline
\textbf{Dataset} & 
  \multicolumn{3}{c|}{\begin{tabular}[c]{@{}c@{}}\textbf{Error in Linear}\\ \textbf{Correlation} $\downarrow$ \end{tabular}} &
  \multicolumn{3}{c|}{\begin{tabular}[c]{@{}c@{}}\textbf{Error in Mutual}\\ \textbf{Information} $\downarrow$ \end{tabular}} \\ \hline
 &
  \multicolumn{1}{c|}{\textbf{MVNet}} &
  \multicolumn{1}{c|}{\textbf{TTHRESH}} &
  \textbf{LERP} &
  \multicolumn{1}{c|}{\textbf{MVNet}} &
  \multicolumn{1}{c|}{\textbf{TTHRESH}} &
  \textbf{LERP} \\ \hline
\textbf{Combustion} &
  \multicolumn{1}{c|}{0.002139} &
  \multicolumn{1}{c|}{0.000919} &
  0.170004 &
  \multicolumn{1}{c|}{0.115754} &
  \multicolumn{1}{c|}{0.226789} &
  0.258035 \\ \hline
\textbf{Isabel} &
  \multicolumn{1}{c|}{0.082532} &
  \multicolumn{1}{c|}{0.188319} &
  0.284613 &
  \multicolumn{1}{c|}{0.195037} &
  \multicolumn{1}{c|}{0.316863} &
  0.227967 \\ \hline
\textbf{Climate50 }&
  \multicolumn{1}{c|}{0.032771} &
  \multicolumn{1}{c|}{0.057524} &
  0.268297 &
  \multicolumn{1}{c|}{0.601893} &
  \multicolumn{1}{c|}{1.535684} &
  0.787852 \\ \hline
\textbf{Climate100} &
  \multicolumn{1}{c|}{0.109252} &
  \multicolumn{1}{c|}{0.137031} &
  0.737098 &
  \multicolumn{1}{c|}{1.554588} &
  \multicolumn{1}{c|}{2.497101} &
  1.963269 \\ \hline
\end{tabular}
}
\end{table}

\begin{table}[tbh]
\centering
\caption{Comparison of the accuracy of multivariate QDV for different methods. Query accuracy between a method and the ground truth result is quantified using the Dice Similarity Coefficient (DSC). It is observed that MVNet produces the most accurate results for both datasets.}
\label{qdv_table}
\resizebox{\linewidth}{!}{
\begin{tabular}{|c|c|c|c|c|}
\hline
\textbf{Dataset} &
  \textbf{\begin{tabular}[c]{@{}c@{}}Multivariate\\ Query\end{tabular}} &
  \textbf{MVNet DSC $\uparrow$} &
  \textbf{TTHRESH DSC $\uparrow$} &
  \textbf{LERP DSC $\uparrow$} \\ \hline
\textbf{Combustion} &
    \begin{tabular}[c]{@{}c@{}}Mixfrac \textgreater 0.3 \& \textless 0.7 \&\\ Y\_OH \textgreater 0.006 \& \textless 0.1\end{tabular} &
  0.9914 &
  0.988 &
  0.8966 \\ \hline
\textbf{Isabel} &
  \begin{tabular}[c]{@{}c@{}}CLOUD \textgreater 0.0001 \& \textless 0.002 \&\\ PRECIP \textgreater 0.0001 \& \textless 0.0093 \&\\ QVAPOR \textgreater 0.01 \& \textless 0.0235\end{tabular} &
  0.7815 &
  0.7083 &
  0.3569 \\ \hline
\end{tabular}}
\end{table}
\subsection{Multivariate Query-Driven Visual Analysis}
\begin{figure}[thb]
\centering
\begin{subfigure}[t]{0.24\linewidth}
    \centering
    \includegraphics[width=\linewidth]{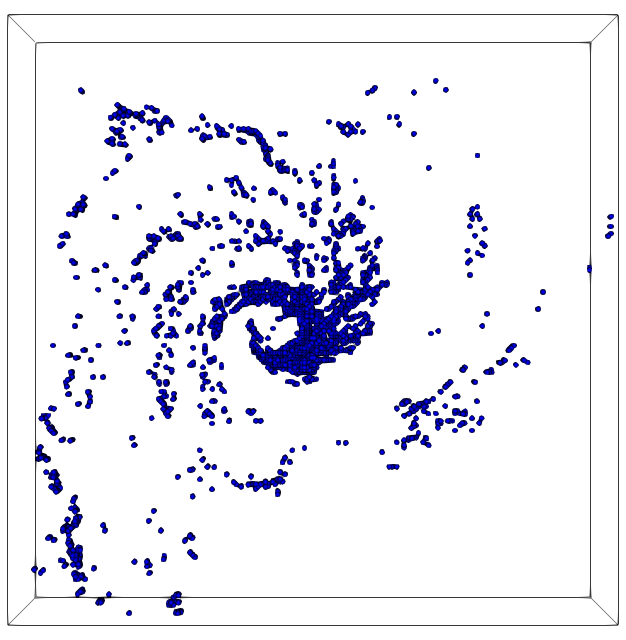}
    \caption{Ground truth}
    \label{isabel_actual_qdv}
\end{subfigure}
~
\begin{subfigure}[t]{0.24\linewidth}
    \centering
     \includegraphics[width=\linewidth]{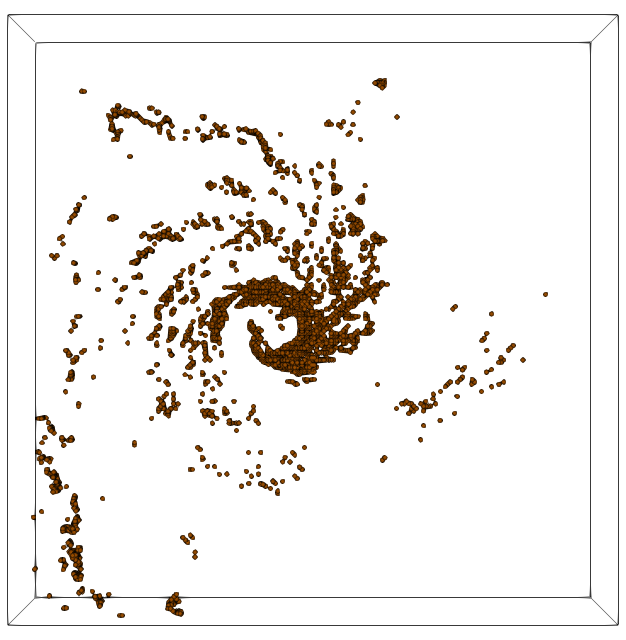}
    \caption{MVNet}
    \label{isabel_mvnet_qdv}
\end{subfigure}
~
\begin{subfigure}[t]{0.24\linewidth}
    \centering
    \includegraphics[width=\linewidth]{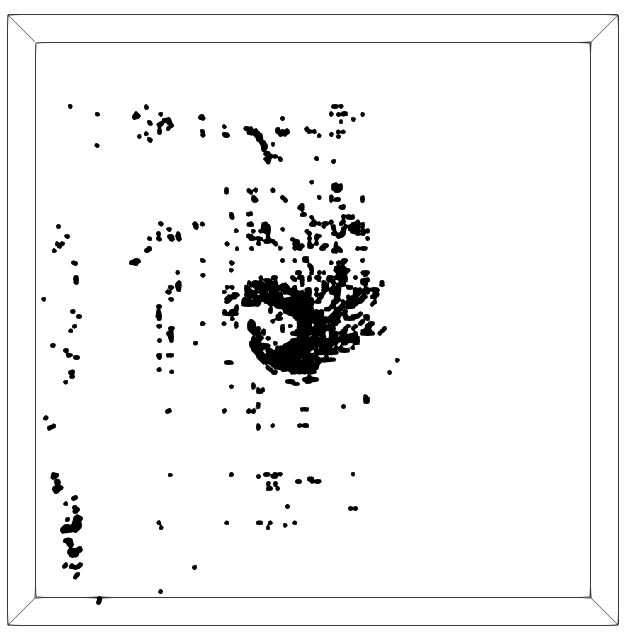}
    \caption{TTHRESH}
    \label{isabel_tthresh_qdv}
\end{subfigure}
~
\begin{subfigure}[t]{0.24\linewidth}
    \centering
    \includegraphics[width=\linewidth]{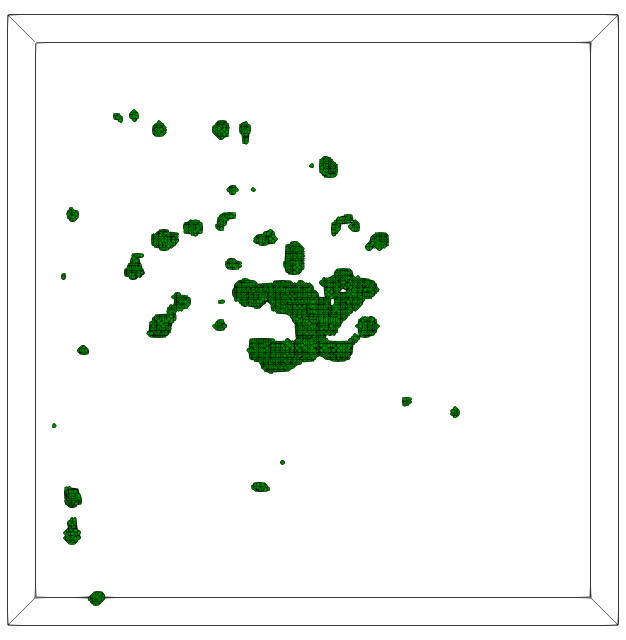}
    \caption{LERP}
    \label{isabel_lerp_qdv}
\end{subfigure}
\caption{Query-driven visualization for the Isabel dataset. A multivariate query: (CLOUD \textgreater 0.0001 AND \textless 0.002) AND (PRECIPITATION \textgreater 0.0001 AND \textless 0.0093) AND (QVAPOR \textgreater 0.01 AND \textless 0.0235) is shown.}
\label{qdv_isabel}
\end{figure}

\begin{figure}[thb]
\centering
\begin{subfigure}[t]{0.24\linewidth}
    \centering
    \includegraphics[width=\linewidth]{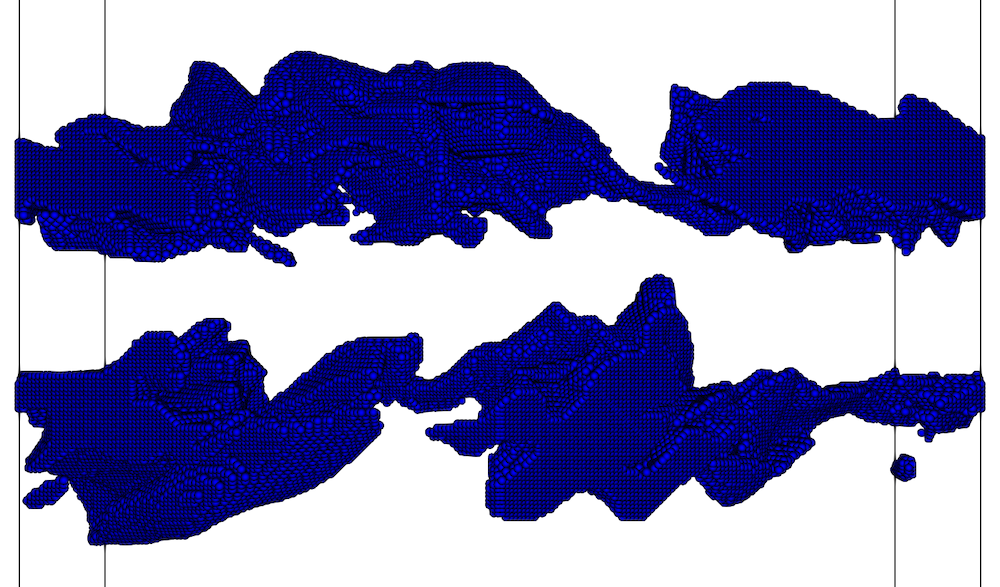}
    \caption{Ground truth}
    \label{comb_gt_qdv}
\end{subfigure}
~
\begin{subfigure}[t]{0.24\linewidth}
    \centering
     \includegraphics[width=\linewidth]{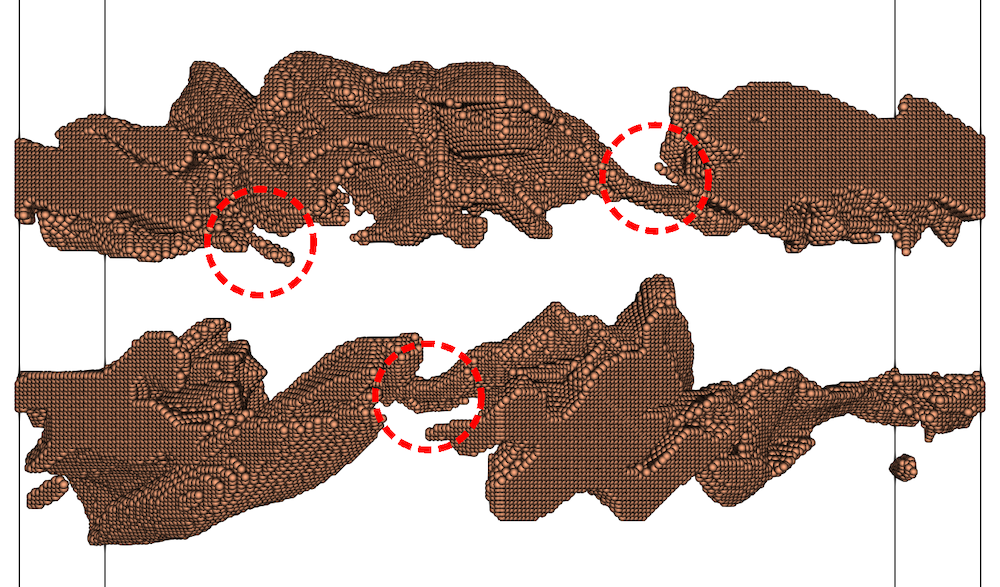}
    \caption{MVNet}
    \label{comb_mvnet_qdv}
\end{subfigure}
~
\begin{subfigure}[t]{0.24\linewidth}
    \centering
    \includegraphics[width=\linewidth]{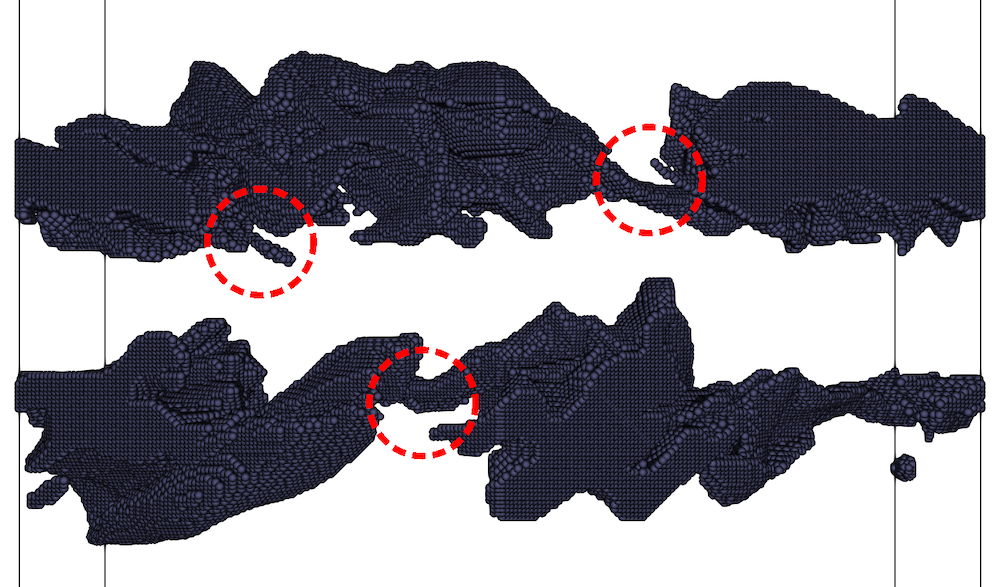}
    \caption{TTHRESH}
    \label{comb_tthresh_qdv}
\end{subfigure}
~
\begin{subfigure}[t]{0.24\linewidth}
    \centering
    \includegraphics[width=\linewidth]{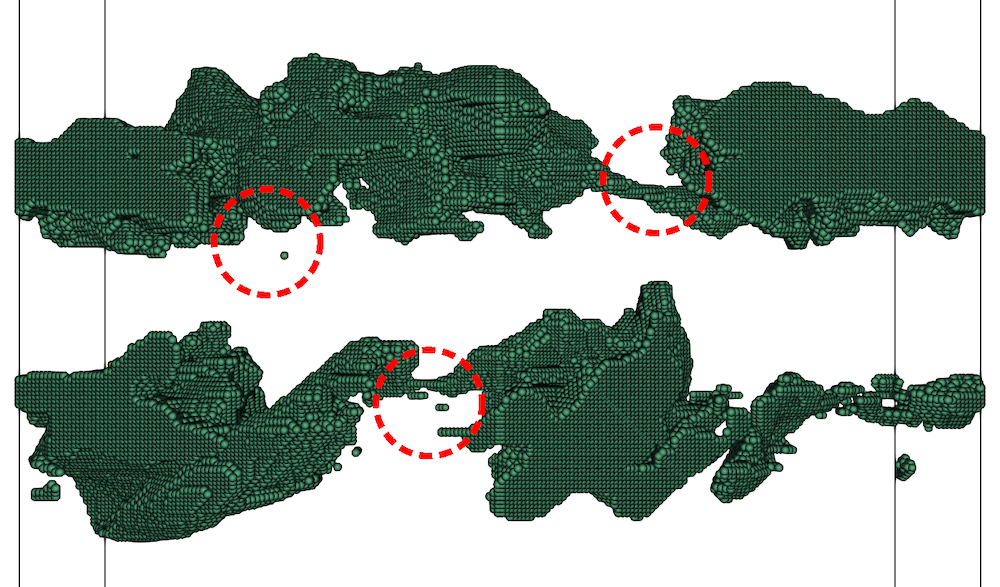}
    \caption{LERP}
    \label{comb_lerp_qdv}
\end{subfigure}
\caption{Query-driven visualization for the Combustion dataset. A multivariate query: (Mixfrac \textgreater 0.3 AND \textless 0.7) AND (Y\_OH \textgreater 0.006 AND \textless 0.1) is shown.}
\label{qdv_comb}
\end{figure}

Multivariate variable interaction analysis using query-driven visualization (QDV) is an effective approach for large scientific data exploration~\cite{qdv1, qdv2, codda, pmi_sampling}. Scientists often perform queries on a group of variables jointly to filter a subset of related data points for detailed feature analysis. Such query-driven analyses help reduce the explorable data size and scientists' cognitive load while accelerating the scientific discovery process. Therefore, we study the efficacy of MVNet in performing multivariate QDV. To compare the accuracy of the QDV results, we compute the Dice Similarity Coefficient (DSC)~\cite{diceIndex} between the query result obtained from ground truth data and the other methods. The value of DSC varies between 0 and 1, where 1 indicates a perfect match. The results and the exact queries are summarized in Table~\ref{qdv_table}.

We execute a query on the Hurricane Isabel dataset involving variables CLOUD, PRECIPITATION, and QVAPOR to study the cloud structure and regions where high rainfall is possible. We jointly query moderate to high value ranges for each variable. The exact query is shown in Table~\ref{qdv_table}. In another experiment, we perform QDV for Combustion data. Turbulent Combustion data is studied to understand the complex combustion process and the interplay among fuel, oxidizer, and other chemical components~\cite{akiba_combustion}. The interaction between mixture fraction (Mixfrac) and mass fraction of Hydroxyl (Y\_OH) variables can be studied to understand the flame structure. Hence, we conduct a QDV using these two variables to isolate data points that indicate the potential flame structure~\cite{akiba_combustion}. From Table~\ref{qdv_table}, we observe that MVNet produces the most accurate query results among all the methods for both datasets. In Fig.~\ref{qdv_isabel} and Fig.~\ref{qdv_comb}, the results of these QDV are shown. We observe that compared to the ground truth, MVNet produces visually similar results, whereas the LERP method produces the least accurate results.

\section{Evaluation \& Parameter Study}
\label{eval_sec}

\begin{figure}[thb]
\centering
\includegraphics[width=0.8\linewidth, height=1.8in]{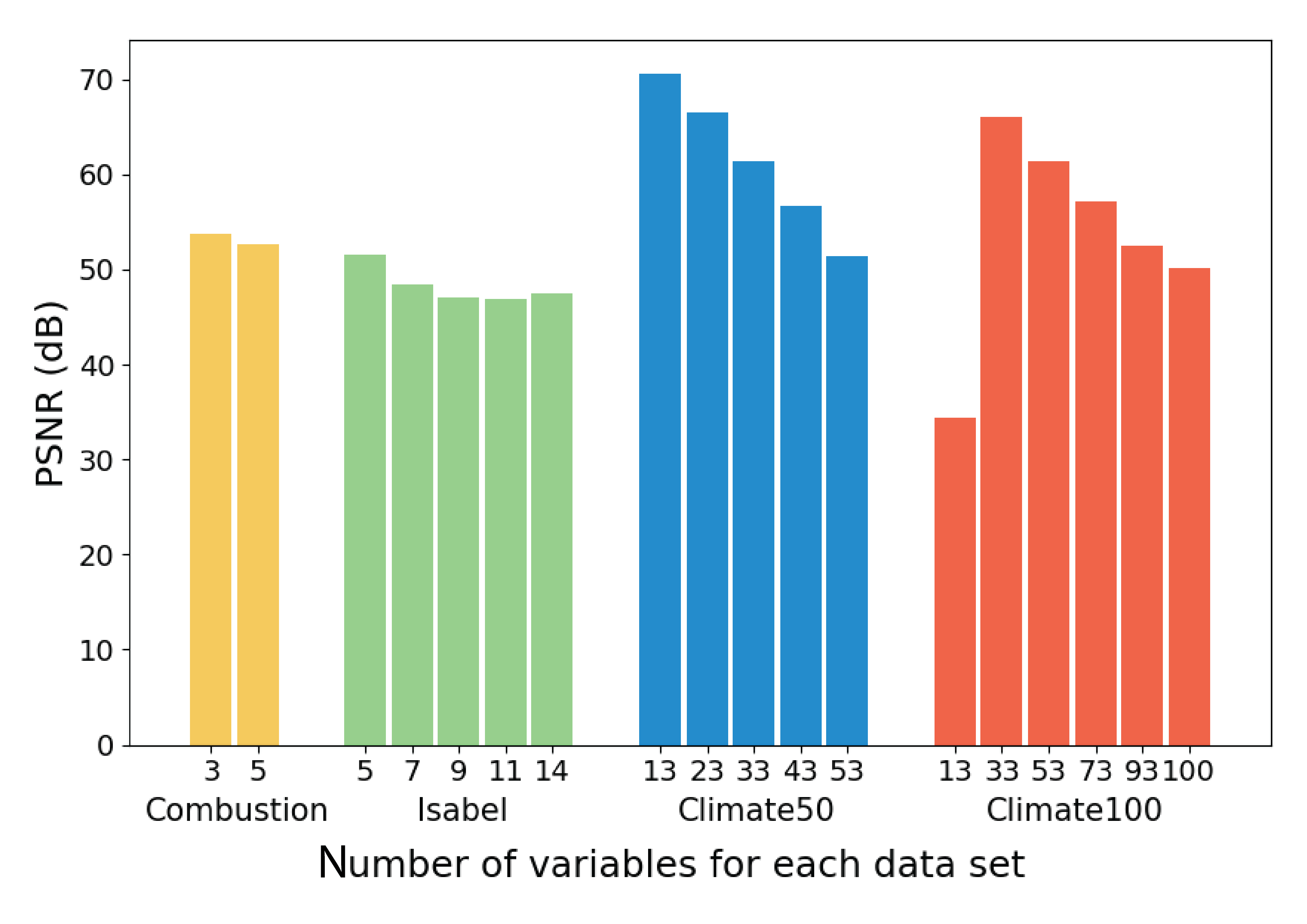}
\caption{The variation of average PSNR when the number of variables increases. We keep the MVNet architecture fixed and only change the number of neurons in the final layer.}
\label{exp_vary_numofvars_barchart}
\end{figure}

\begin{figure}[thb]
\centering
\includegraphics[width=0.85\linewidth]{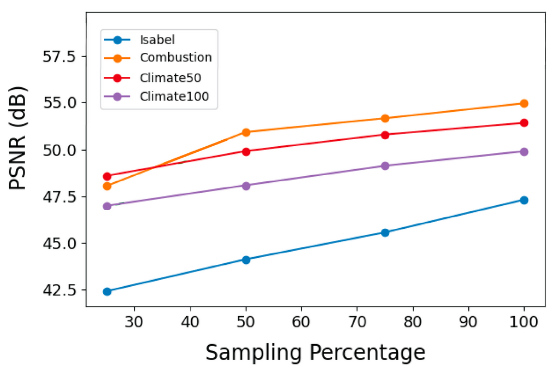}
\caption{Average PSNR values under different sampling percentages for different datasets.}
\label{exp_vary_samplingPercent}
\end{figure}

\subsection{Varying Number of Variables}
To evaluate the performance of MVNet in this regard, we conduct experiments where we gradually increase the number of variables while keeping the architecture of MVNet fixed. Only the number of neurons in the output layer is adjusted to accommodate the varying number of variables. All other hyperparameters are set to the best configuration determined from experimentation (details are in supplementary document). The results are depicted in Fig.~\ref{exp_vary_numofvars_barchart} where different colors representing different datasets. As the number of variables increases, the learning task becomes more challenging, leading to a gradual decrease in the average PSNR value across all variables. However, it's worth noting that even when trained with the maximum number of variables, MVNet maintains a high average PSNR for each dataset. Furthermore, Table~\ref{vary_resblocks_table} highlights that employing a deeper MVNet model can further enhance PSNR, providing the option for users to obtain higher-quality reconstructions if desired. These findings offer valuable insights for users regarding the trade-off between expected quality and storage overhead across different datasets and variable configurations.

\begin{table*}[tb!]
\centering
\caption{Average PSNR, storage footprint, and compression ratio (CR) under varying numbers of residual blocks to study the trade-off between compression ratio and PSNR. We observe that MVNet maintains a high PSNR even when fewer residual blocks are used.}
\label{vary_resblocks_table}
\resizebox{\linewidth}{!}{
\begin{tabular}{|c|ccc|ccc|ccc|ccc|ccc|ccc|}
\hline
\multirow{2}{*}{\textbf{Dataset}} &
  \multicolumn{3}{c|}{\textbf{Num. of Res. Block = 4}} &
  \multicolumn{3}{c|}{\textbf{Num. of Res. Block = 6}} &
  \multicolumn{3}{c|}{\textbf{Num. of Res. Block = 8}} &
  \multicolumn{3}{c|}{\textbf{Num. of Res. Block = 10}} &
  \multicolumn{3}{c|}{\textbf{Num. of Res. Block = 12}} &
  \multicolumn{3}{c|}{\textbf{Num. of Res. Block = 14}} \\ \cline{2-19} 
 &
  \multicolumn{1}{c|}{\textbf{\begin{tabular}[c]{@{}c@{}}Storage\\ (KB)\end{tabular}}} &
  \multicolumn{1}{c|}{\textbf{\begin{tabular}[c]{@{}c@{}}CR\end{tabular}}} &
  \textbf{\begin{tabular}[c]{@{}c@{}}PSNR\\ (dB)\end{tabular}} &
  \multicolumn{1}{c|}{\textbf{\begin{tabular}[c]{@{}c@{}}Storage\\ (KB)\end{tabular}}} &
  \multicolumn{1}{c|}{\textbf{\begin{tabular}[c]{@{}c@{}}CR\end{tabular}}} &
  \textbf{\begin{tabular}[c]{@{}c@{}}PSNR\\ (dB)\end{tabular}} &
  \multicolumn{1}{c|}{\textbf{\begin{tabular}[c]{@{}c@{}}Storage\\ (KB)\end{tabular}}} &
  \multicolumn{1}{c|}{\textbf{\begin{tabular}[c]{@{}c@{}}CR\end{tabular}}} &
  \textbf{\begin{tabular}[c]{@{}c@{}}PSNR\\ (dB)\end{tabular}} &
  \multicolumn{1}{c|}{\textbf{\begin{tabular}[c]{@{}c@{}}Storage\\ (KB)\end{tabular}}} &
  \multicolumn{1}{c|}{\textbf{\begin{tabular}[c]{@{}c@{}}CR\end{tabular}}} &
  \textbf{\begin{tabular}[c]{@{}c@{}}PSNR\\ (dB)\end{tabular}} &
  \multicolumn{1}{c|}{\textbf{\begin{tabular}[c]{@{}c@{}}Storage\\ (KB)\end{tabular}}} &
  \multicolumn{1}{c|}{\textbf{\begin{tabular}[c]{@{}c@{}}CR\end{tabular}}} &
  \textbf{\begin{tabular}[c]{@{}c@{}}PSNR\\ (dB)\end{tabular}} &
  \multicolumn{1}{c|}{\textbf{\begin{tabular}[c]{@{}c@{}}Storage\\ (KB)\end{tabular}}} &
  \multicolumn{1}{c|}{\textbf{\begin{tabular}[c]{@{}c@{}}CR\end{tabular}}} &
  \textbf{\begin{tabular}[c]{@{}c@{}}PSNR\\ (dB)\end{tabular}} \\ \hline
\textbf{Combustion} &
  \multicolumn{1}{c|}{468} &
  \multicolumn{1}{c|}{236.2:1} &
  45.567 &
  \multicolumn{1}{c|}{700} &
  \multicolumn{1}{c|}{157.9:1} &
  48.386 &
  \multicolumn{1}{c|}{928} &
  \multicolumn{1}{c|}{119.14:1} &
  50.701 &
  \multicolumn{1}{c|}{1160} &
  \multicolumn{1}{c|}{95.3:1} &
  52.521 &
  \multicolumn{1}{c|}{1388} &
  \multicolumn{1}{c|}{79.65:1} &
  53.984 &
  \multicolumn{1}{c|}{1620} &
  \multicolumn{1}{c|}{68.25:1} &
  55.411 \\ \hline
\textbf{Isabel} &
  \multicolumn{1}{c|}{472} &
  \multicolumn{1}{c|}{343.13:1} &
  44.207 &
  \multicolumn{1}{c|}{704} &
  \multicolumn{1}{c|}{230.05:1} &
  45.464 &
  \multicolumn{1}{c|}{932} &
  \multicolumn{1}{c|}{173.77:1} &
  46.381 &
  \multicolumn{1}{c|}{1164} &
  \multicolumn{1}{c|}{139.13:1} &
  47.634 &
  \multicolumn{1}{c|}{1396} &
  \multicolumn{1}{c|}{116.01:1} &
  47.947 &
  \multicolumn{1}{c|}{1624} &
  \multicolumn{1}{c|}{99.72:1} &
  49.168 \\ \hline
\textbf{Climate50} &
  \multicolumn{1}{c|}{492} &
  \multicolumn{1}{c|}{1753.38:1} &
  44.038 &
  \multicolumn{1}{c|}{720} &
  \multicolumn{1}{c|}{1198.14:1} &
  46.697 &
  \multicolumn{1}{c|}{952} &
  \multicolumn{1}{c|}{906.15:1} &
  48.740 &
  \multicolumn{1}{c|}{1184} &
  \multicolumn{1}{c|}{728.6:1} &
  51.367 &
  \multicolumn{1}{c|}{1412} &
  \multicolumn{1}{c|}{610.95:1} &
  52.805 &
  \multicolumn{1}{c|}{1644} &
  \multicolumn{1}{c|}{524.73:1} &
  54.152 \\ \hline
\textbf{Climate100 }&
  \multicolumn{1}{c|}{512} &
  \multicolumn{1}{c|}{3172.01:1} &
  42.784 &
  \multicolumn{1}{c|}{744} &
  \multicolumn{1}{c|}{2182.89:1} &
  45.303 &
  \multicolumn{1}{c|}{976} &
  \multicolumn{1}{c|}{1664:1} &
  47.380 &
  \multicolumn{1}{c|}{1204} &
  \multicolumn{1}{c|}{1348.89:1} &
  49.991 &
  \multicolumn{1}{c|}{1436} &
  \multicolumn{1}{c|}{1130.96:1} &
  51.515 &
  \multicolumn{1}{c|}{1664} &
  \multicolumn{1}{c|}{976:1} &
  52.624 \\ \hline
\end{tabular}
}\end{table*}

\begin{table}[tbh]
\caption{Hyperparameters, training and inference time. All experiments are done on a GPU server with NVIDIA GeForce GTX 1080Ti GPUs with 12GB GPU memory}
\label{train_inf_table}
\resizebox{\linewidth}{!}{%
\begin{tabular}{|c|c|c|c|c|c|c|}
\hline
\textbf{Dataset} &
  \textbf{\begin{tabular}[c]{@{}c@{}}Learning\\ Rate\end{tabular}} &
  \textbf{\begin{tabular}[c]{@{}c@{}}Batch \\ Size\end{tabular}} &
  \textbf{\begin{tabular}[c]{@{}c@{}}Decay \\ Rate\end{tabular}} &
  \textbf{\begin{tabular}[c]{@{}c@{}}Decay \\ Frequency\end{tabular}} &
  \textbf{\begin{tabular}[c]{@{}c@{}}Training \\ Time (Hrs.)\end{tabular}} &
  \textbf{\begin{tabular}[c]{@{}c@{}}Inference \\ Time (Secs.)\end{tabular}} \\ \hline
\textbf{Combustion} & 0.00005 & 2048 & 0.8 & Every 15th Epoch & 3.89 & 2.04  \\ \hline
\textbf{Isabel}     & 0.0001  & 2048 & 0.8 & Every 15th Epoch & 2.32 & 1.46  \\ \hline
\textbf{Climate50}  & 0.0001  & 2048 & 0.8 & Every 15th Epoch & 2.72 & 3.96  \\ \hline
\textbf{Climate100} & 0.0001  & 2048 & 0.8 & Every 15th Epoch & 2.93 & 10.38 \\ \hline
\end{tabular}%
}
\end{table}

\begin{figure}[thb]
\centering
\includegraphics[width=0.9\linewidth]{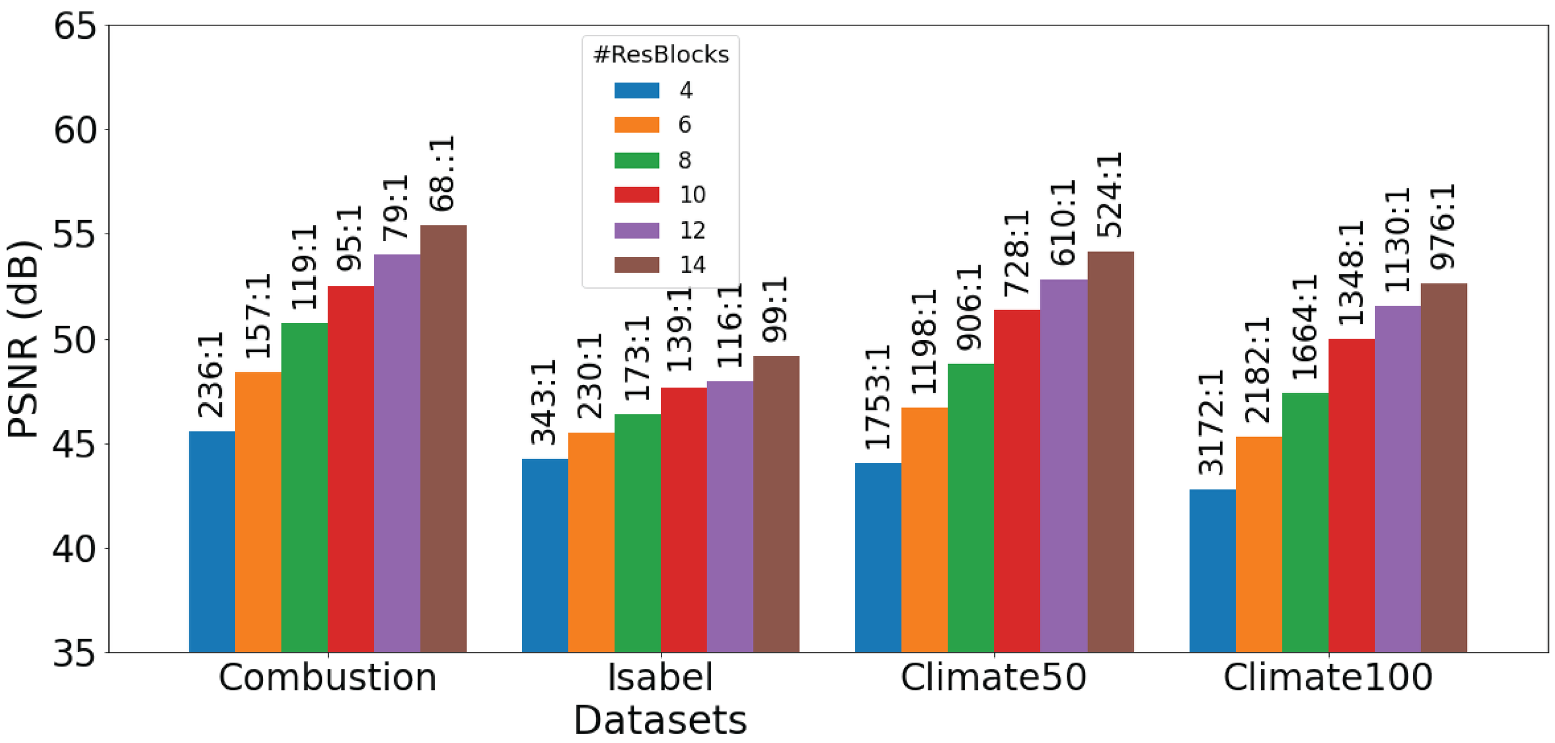}
\caption{Change of average PSNR when number of res. blocks are varied. The corresponding compression ratio is also shown at the top of each bar.}
\label{PSNR_Compression}
\end{figure}

\subsection{Compression Ratio vs. Prediction Quality}
\label{vary_resblocks}
The architecture of MVNet relies on residual blocks, which allows for increasing the network's depth to potentially enhance prediction quality and generalizability. However, this also increases model complexity, leading to higher storage requirements. To understand the trade-off between storage and prediction quality, we conducted an experiment varying the number of residual blocks from $4$ to $14$, while keeping the number of neurons per hidden layer constant at $120$. This analysis also sheds light on MVNet's compression capabilities. The results, presented in Table~\ref{vary_resblocks_table}, reveal that with only $4$ residual blocks, MVNet achieves maximum compression ratio while maintaining good average PSNR across datasets. This trade-dff is also provided as a form of grouped bar chart in Fig.~\ref{PSNR_Compression}. Notably, as the number of variables increases in multivariate datasets, MVNet offers higher compression ratios due to its consistent network size.

\subsection{Varying Amounts of Training Data}
As computational capabilities advance, storing entire high-resolution multivariate datasets becomes increasingly costly. In such cases, storing sub-sampled data becomes a viable option to reduce storage requirements for offline analysis and visualization. Consequently, training MVNet on sub-sampled data becomes necessary. To investigate the effectiveness of MVNet under these circumstances, we conducted a comprehensive experiment by training MVNet using 25\%, 50\%, 75\%, and 100\% data samples, randomly selected uniformly. The Fig.~\ref{exp_vary_samplingPercent} illustrates the average quality of the reconstructed data variables for various datasets. We observe a steady increase in reconstruction quality with more training samples. However, it's noteworthy that the rate of improvement in PSNR values isn't significantly high with increasing sample points. Even at a low sampling rate of 25\%, MVNet can still achieve very good average PSNR, suggesting its efficacy when trained on sub-sampled data.

\subsection{Hyperparameters, Training, and Inference Timing of MVNet}
The hyperparameters, timings for both training and inference of MVNet are presented in Table~\ref{train_inf_table}. Consistency is maintained by training MVNet for 300 epochs across all datasets. It's demonstrated that MVNet achieves rapid reconstruction (inference) of all variables, with inference time correlating to the number of variables in the multivariate dataset. Table~\ref{train_inf_table} illustrates training time using the optimal hyperparameter combinations for each dataset. Through comprehensive evaluation, it's noted that increasing the batch size decreases training time, and training time scales with both the number of data points and variables.

\section{Discussion}

This work demonstrates the effectiveness of utilizing residual implicit neural representations with sinusoidal activation functions to model complex multivariate scientific datasets when initialized with carefully selected weights and hyperparameters. Our primary focus is on datasets featuring a large number of variables with dynamic characteristics, enabling us to assess the model's performance comprehensively. We compare MVNet against established methods such as tensor compressor TTHRESH, Zfp, and Copula modeling, and employ various metrics to assess reconstructed data, feature, and image quality. We further delve into MVNet's ability to recover both linear and nonlinear dependencies among variable combinations, which is important for multivariate data analysis. We then comparatively study the efficacy of MVNet in carrying out important multivariate tasks such as query-driven visual analysis. We conduct experiments to report the trade-off between compression ratio and prediction quality, allowing users to use this as a guideline for selecting the appropriate model architecture for their dataset. We also depict how increasing variables in the dataset impact MVNet's overall prediction quality. Depending on the need for reconstruction quality and the number of variables to be modeled, the users can select an appropriate depth for MVNet to get a desirable accuracy. Through our thorough evaluation, we aim to encourage domain scientists to integrate INR-based models, such as MVNet, into their analysis workflows. These models provide a compact yet versatile representation of multivariate data, facilitating streamlined analytics and visualization, and accelerating the scientific discovery process.

\section{Conclusions and Future Work}

In summary, we introduce MVNet to efficiently capture complex multivariate scientific data containing a large number of variables. We undertake a comprehensive investigation to showcase the effectiveness of MVNet compared to state-of-the-art compression and statistical methods, focusing on the balance between storage and the reconstruction quality, as well as its performance in several scientific applications. Looking ahead, we aim to expand this work by applying MVNet to model time-varying scientific datasets including spatio-temporal vector and tensor data.

\bibliographystyle{ACM-Reference-Format}
\bibliography{template}

\newpage
\appendix
\section{Supplementary Material}

\subsection{Comparison of Gradient Quality}
\begin{figure*}[!t]
\centering
\begin{subfigure}[t]{0.22\linewidth}
    \centering
    \includegraphics[width=\linewidth]{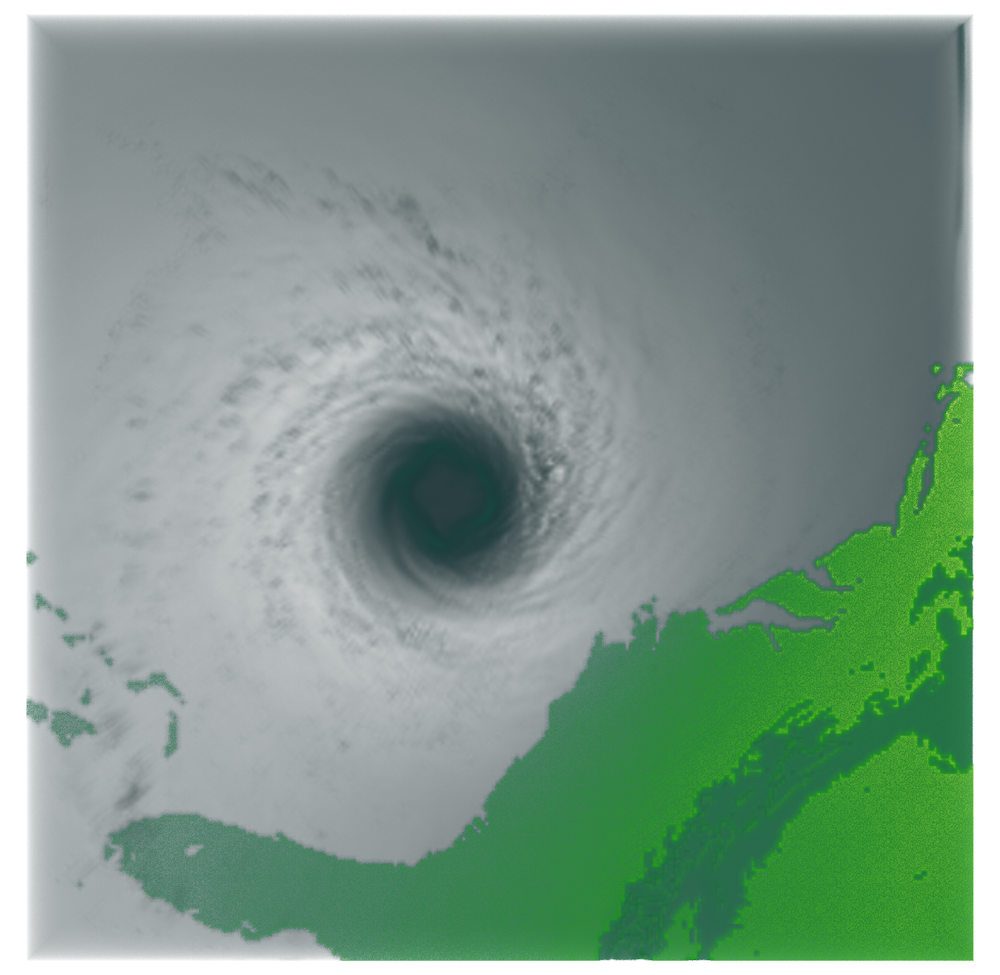}
    \caption{Ground truth grad. mag. of P variable.}
    \label{isabel_grad_P_GT}
\end{subfigure}
~
\begin{subfigure}[t]{0.22\linewidth}
    \centering
    \includegraphics[width=\linewidth]{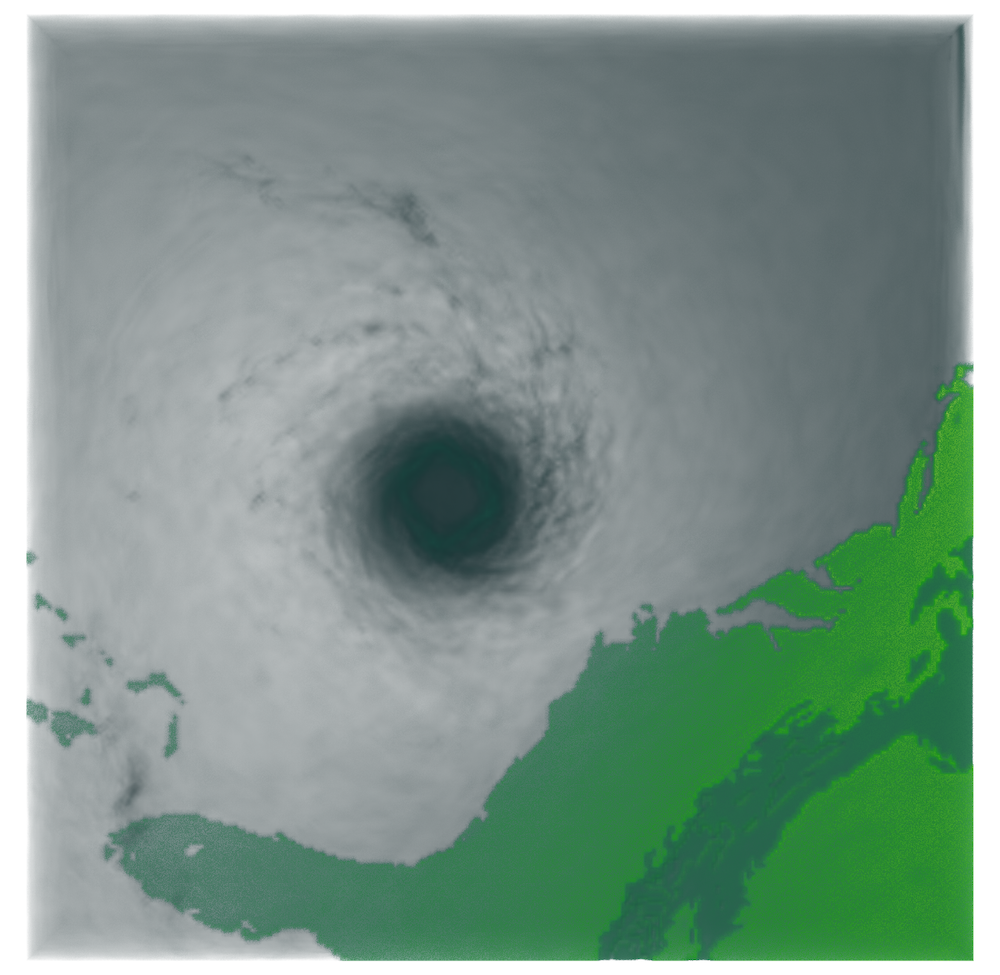}
    \caption{MVNet generated grad. mag. of P variable.}
    \label{isabel_grad_P_MVNet}
\end{subfigure}
~
\begin{subfigure}[t]{0.22\linewidth}
    \centering
    \includegraphics[width=\linewidth]{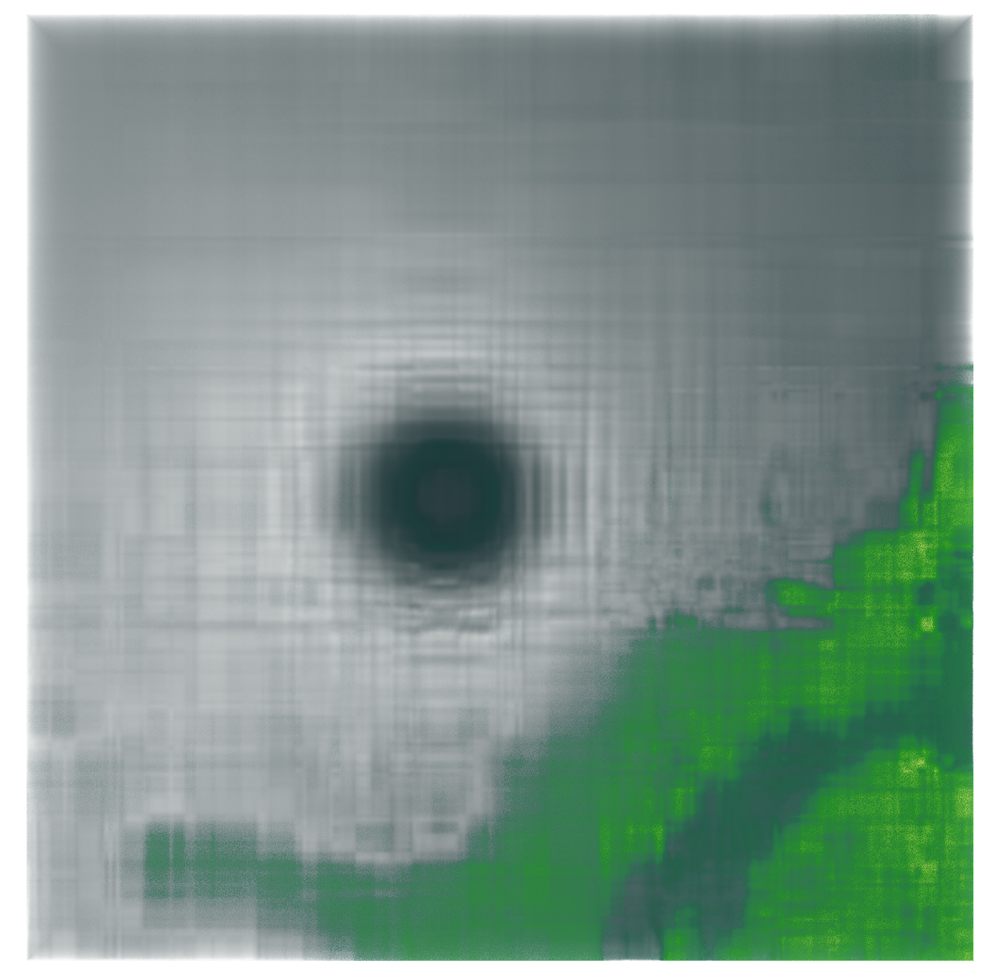}
    \caption{TTHRESH generated grad. mag. of P variable.}
    \label{isabel_grad_P_TTHRESH}
\end{subfigure}
~
\begin{subfigure}[t]{0.22\linewidth}
    \centering
    \includegraphics[width=\linewidth]{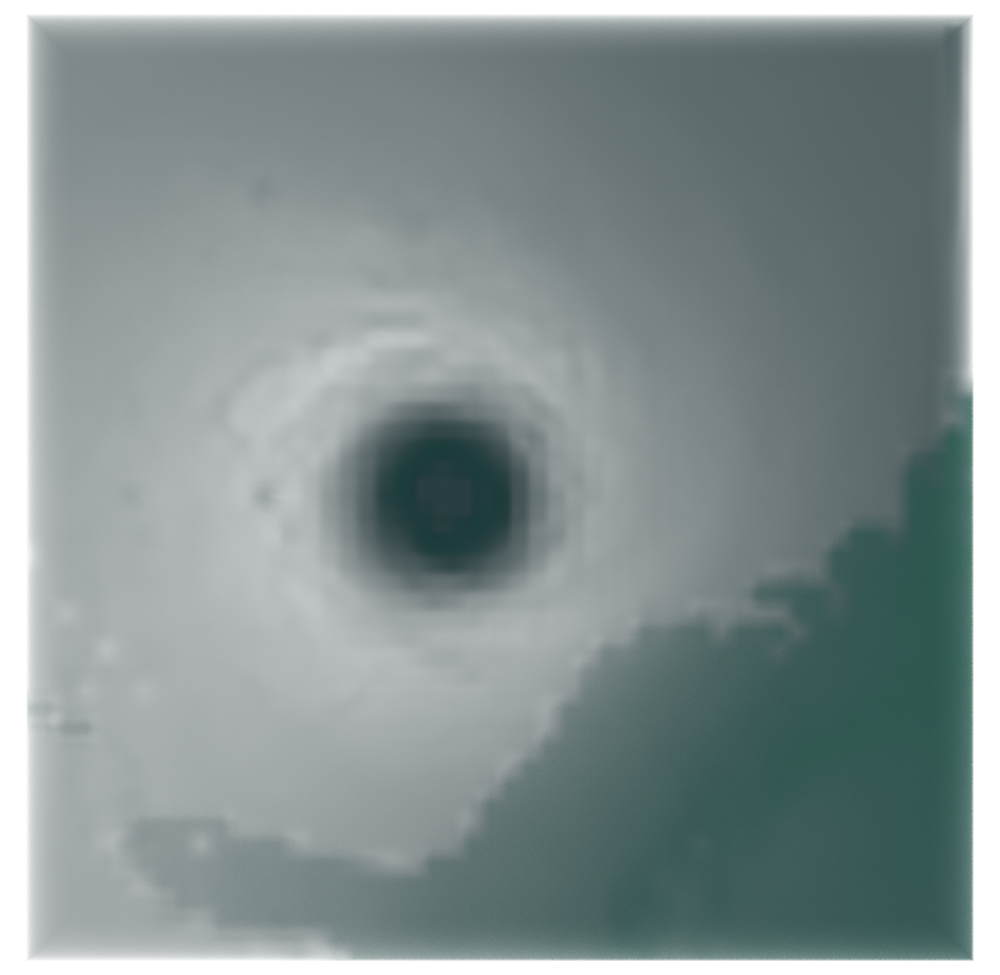}
    \caption{LERP generated grad. mag. of P variable.}
    \label{isabel_grad_P_LERP}
\end{subfigure}
\caption{Reconstructed gradient magnitude visualization for Pressure (P) variable of Isabel dataset. From left to right, ground truth, MVNet-generated result, TTHRESH-generated result, and LERP-generated result are provided. We observe that MVNet produces the most accurate gradient reconstruction, while the TTHRESH and LERP-generated images have artifacts, resulting in reduced accuracy.}
\label{isabel_vis_grad}
\end{figure*}

The gradient of scientific data variables has been shown to be an important quantity to carry out several data analysis tasks, such as importance-driven sampling and multi-dimensional transfer function design. Therefore, we conduct a comparison based on the quality of the gradient computed per grid point from the reconstructed volume data among the three methods. After reconstructing all the variables, we compute spatial gradient magnitudes for each grid location and then compute the PSNR value between the gradient magnitudes estimated from the ground truth data and the gradients from the reconstructed data. Table~\ref{gradient_table} shows the average PNSR values for all the variables for the three methods. It is observed that the proposed MVNet results in the most accurate gradients using the reconstructed data variable values for all the datasets. In Fig.~\ref{isabel_vis_grad}, we show the visualization of the reconstructed gradient magnitude for a representative variable, Pressure (P), of the Isabel dataset. We observe that the TTHRESH (Fig.~\ref{isabel_grad_P_TTHRESH}) and LERP (Fig.~\ref{isabel_grad_P_LERP}) generated gradient fields show visual artifacts, while MVNet (Fig.~\ref{isabel_grad_P_MVNet}) produces the most accurate gradient reconstruction when compared against ground truth (Fig.~\ref{isabel_grad_P_GT}).

\subsection{Comparison with Copula-based Method: Visualization Result}
\begin{figure*}[thb]
\centering
\begin{subfigure}[t]{0.32\linewidth}
    \centering
    \includegraphics[width=\linewidth]{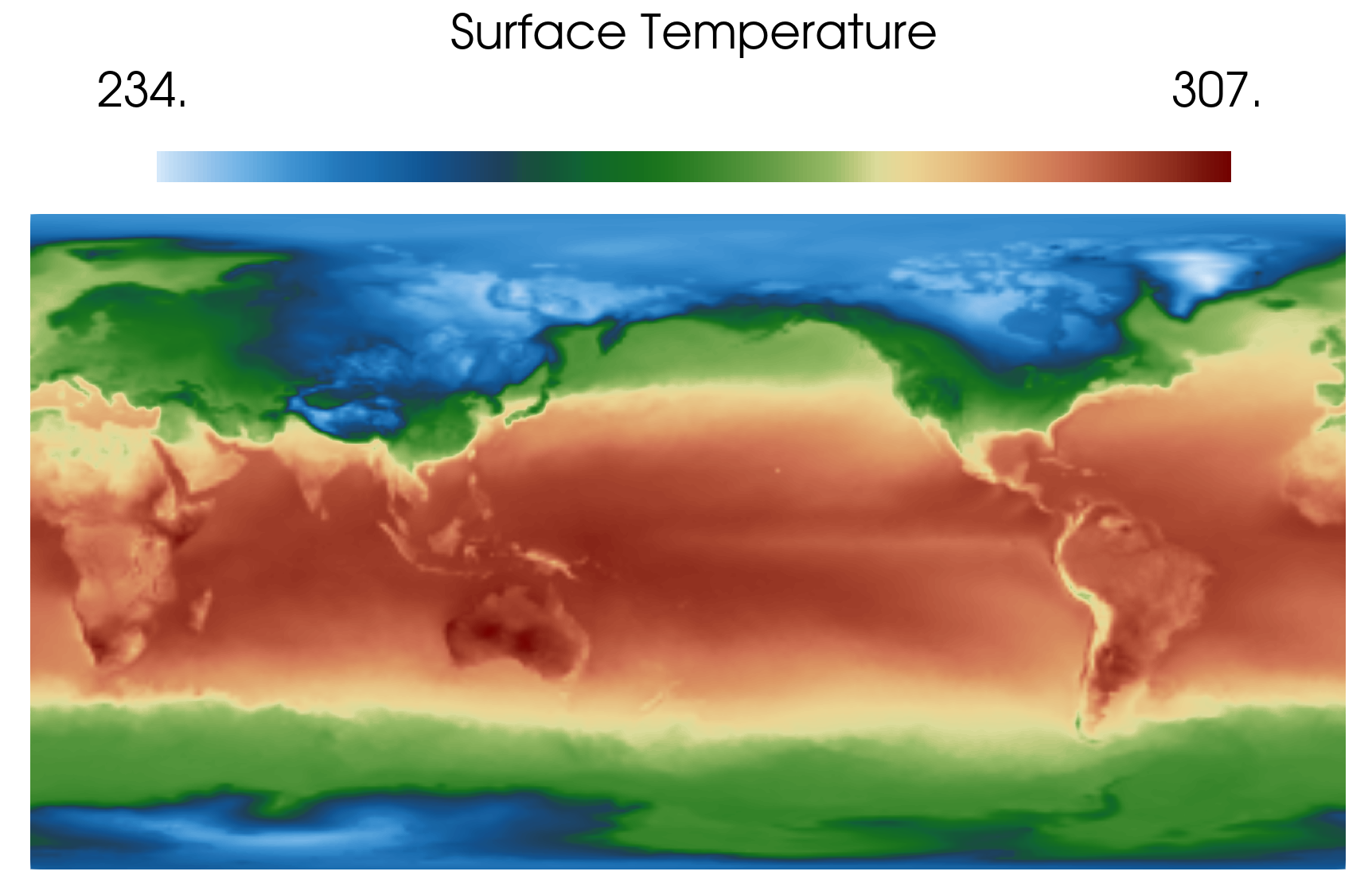}
    \caption{Ground truth visualization of TS variable.}
    \label{GT_TS_climate100}
\end{subfigure}
~
\begin{subfigure}[t]{0.32\linewidth}
    \centering
    \includegraphics[width=\linewidth]{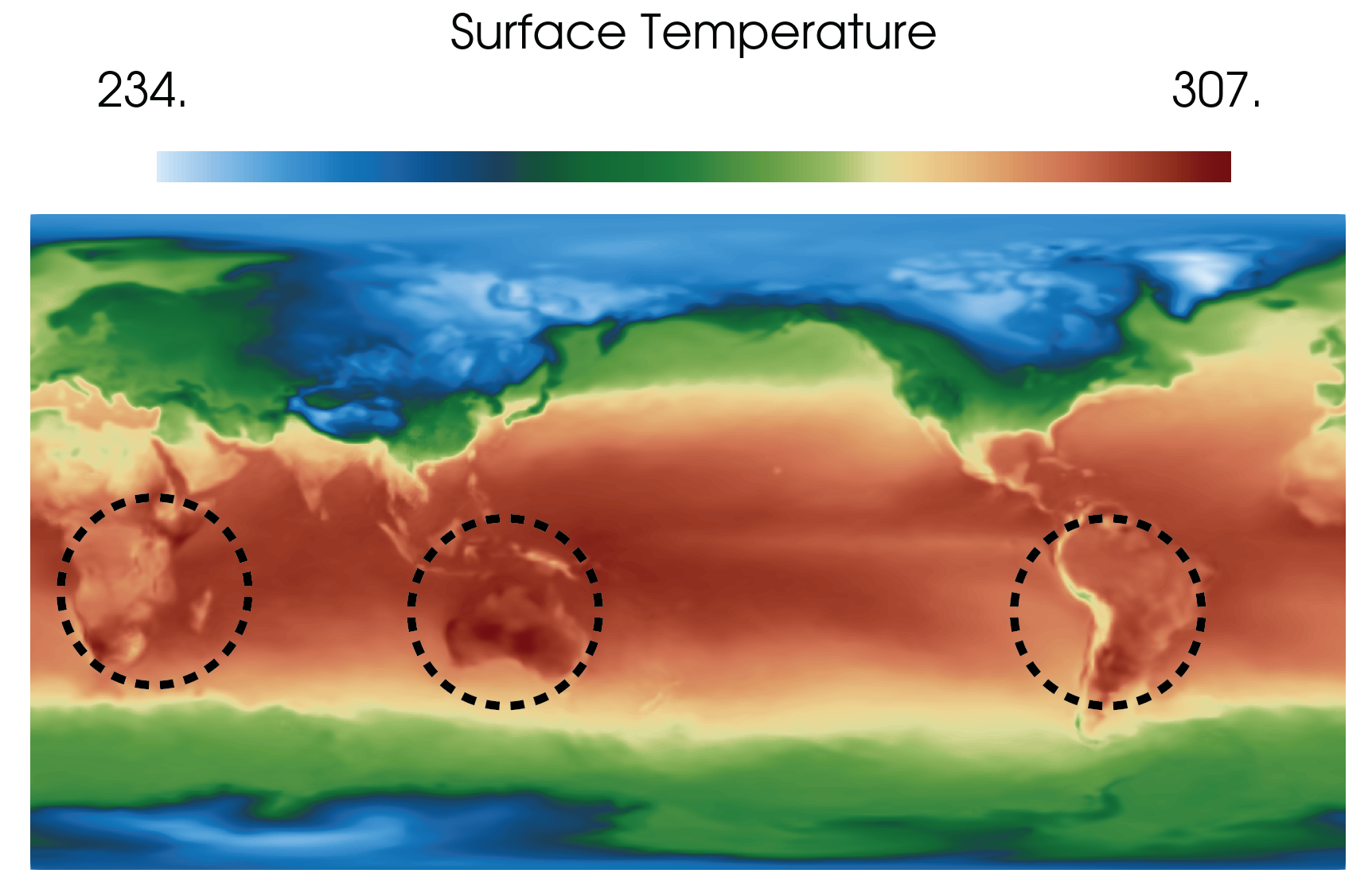}
    \caption{MVNet generated visualization of TS variable.}
    \label{mvnet_TS_climate100}
\end{subfigure}
~
\begin{subfigure}[t]{0.32\linewidth}
    \centering
    \includegraphics[width=\linewidth]{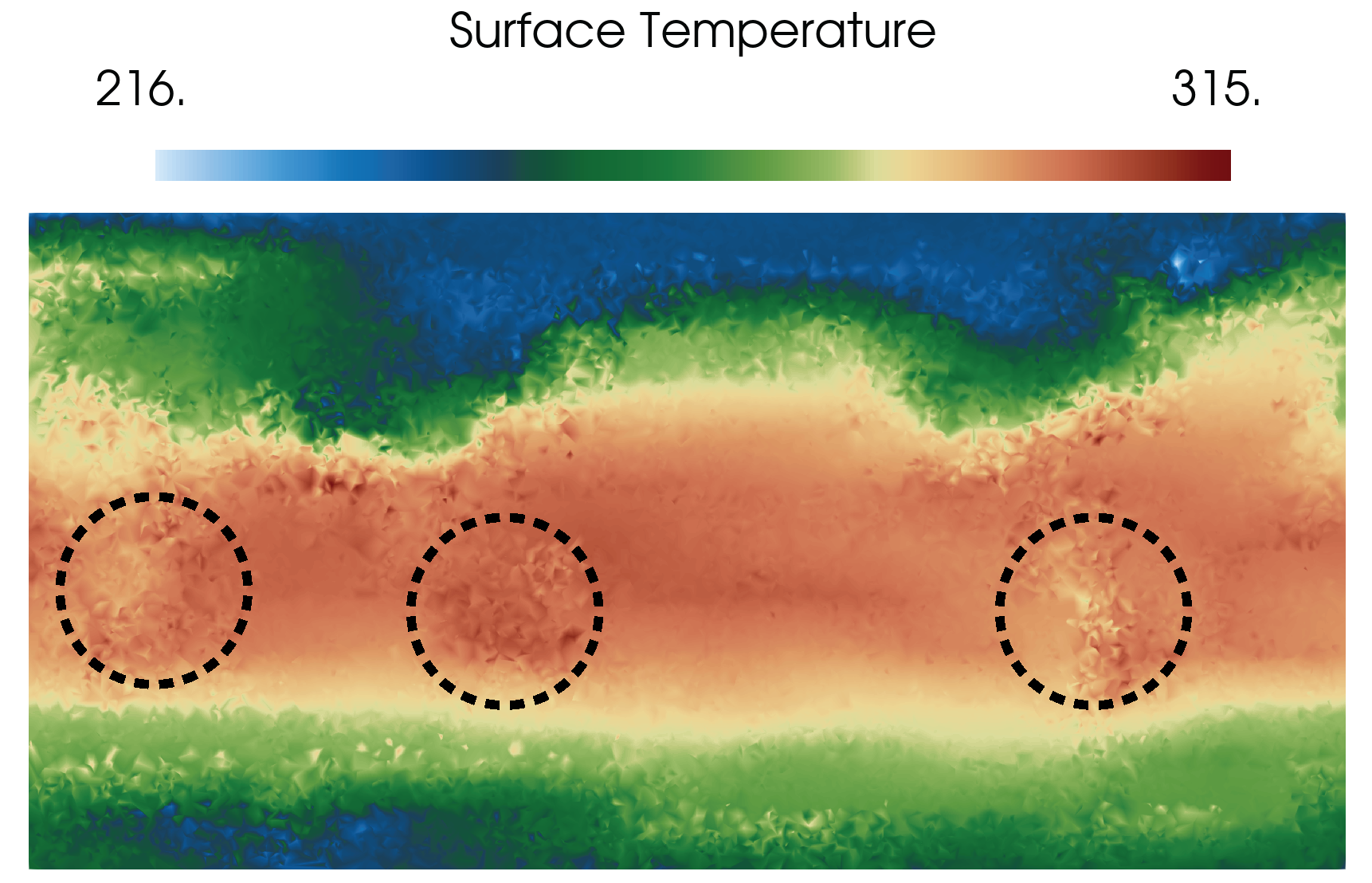}
    \caption{Copula-based summary generated visualization of TS variable.}
    \label{copula_TS_climate100}
\end{subfigure}
\caption{Visualization of Surface Temperature (TS) as a representative variable from the Climate100 dataset. Fig.~\ref{GT_TS_climate100} shows the ground truth, Fig.~\ref{mvnet_TS_climate100} shows the result reconstructed by MVNet, and Fig.~\ref{copula_TS_climate100} depicts the visualization reconstructed using the Copula-based summary data. It is observed that MVNet accurately reconstructs the TS variable. In contrast, the copula-based method fails to preserve the detailed features in the data, as highlighted by black dotted circles in the image.}
\label{copula_vis}
\end{figure*}

In order to compare the quality of the reconstructed data variables visually, we show a rendering of a representative variable, Surface Temperature (TS), using MVNet and Copula-based summaries in Fig.~\ref{copula_vis}. It is seen that MVNet produces a visualization (Fig.~\ref{mvnet_TS_climate100}) identical to the ground truth (Fig.~\ref{GT_TS_climate100}). In contrast, the visualization generated by Copula-based summaries (Fig.~\ref{copula_TS_climate100}) fails to preserve the detailed features of the TS variable as indicated by black dotted circles in the image.

\begin{table}[thb]
\centering
\caption{Quantitative comparison of average gradient magnitude values for all the variables between MVNet and other methods. It is observed that the PSNR (dB) value computed between raw data gradient and reconstructed data gradient is highest for MVNet compared to TTHRESH and LERP.}
\label{gradient_table}
\resizebox{0.9\linewidth}{!}{
\begin{tabular}{|c|c|c|c|}
\hline
\textbf{Dataset}     & \multicolumn{1}{c|}{\textbf{MVNet PSNR $\uparrow$}} & \multicolumn{1}{c|}{\textbf{TTHRESH PSNR $\uparrow$}} & \multicolumn{1}{c|}{\textbf{LERP PSNR $\uparrow$}} \\ \hline
\textbf{ Combustion }& 52.930 & 49.566 & 35.179 \\ \hline
\textbf{ Isabel }    & 46.732 & 45.204 & 39.460 \\ \hline
\textbf{Climate50  }         & 38.453 & 31.909 & 38.015 \\ \hline
\textbf{Climate100 } & 38.911 & 33.896 & 38.363 \\ \hline
\end{tabular}
}
\end{table}

\begin{figure*}[thb]
\centering
\begin{subfigure}[t]{0.24\linewidth}
    \centering
    \includegraphics[width=\linewidth]{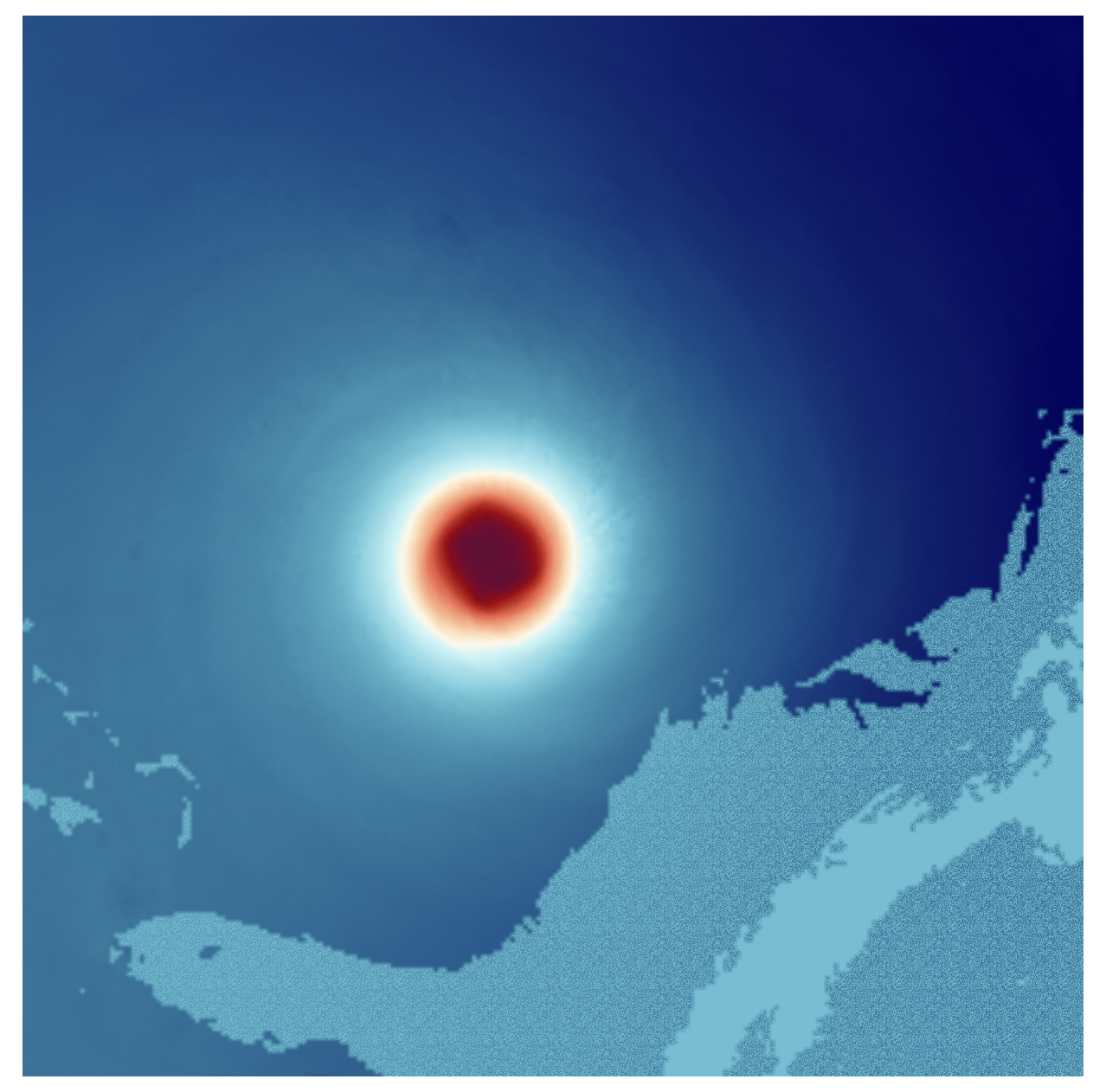}
    \caption{\raggedright GT}

    \label{Isabel_P_GT}
\end{subfigure}
~
\begin{subfigure}[t]{0.24\linewidth}
    \centering
    \includegraphics[width=\linewidth]{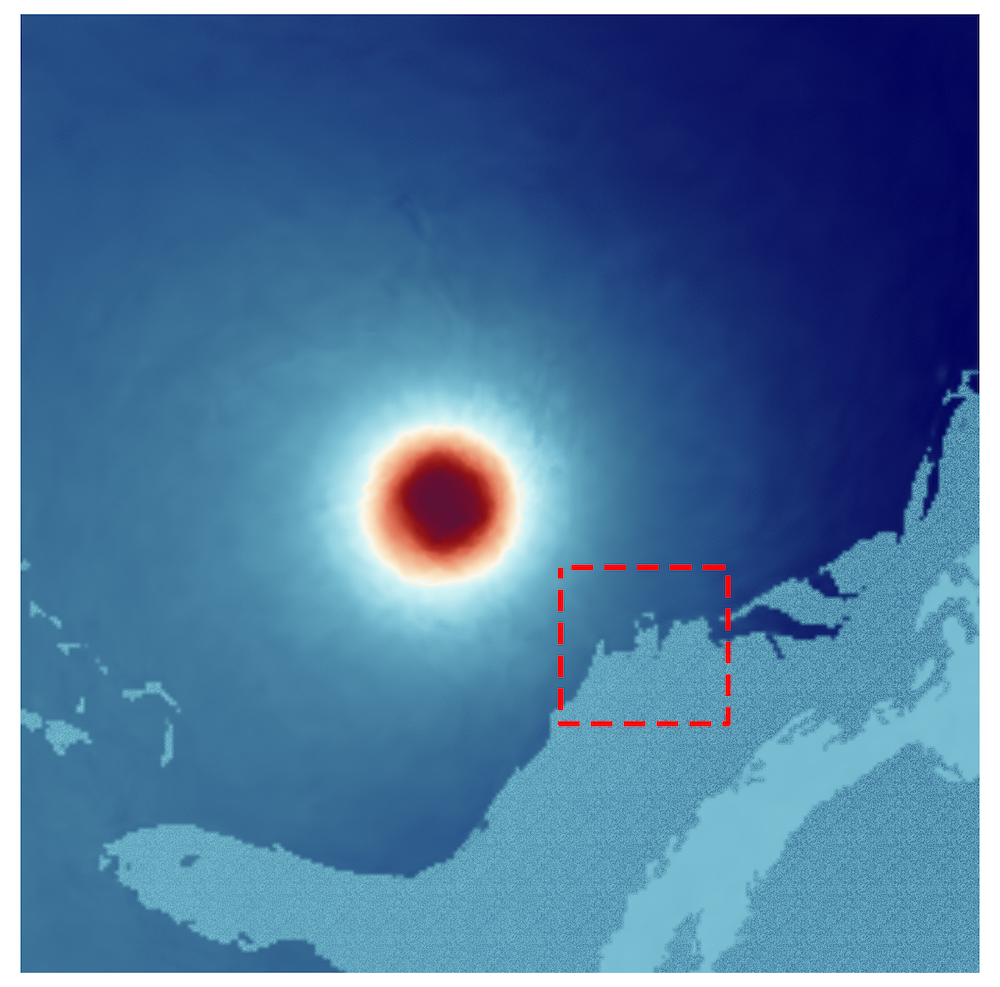}
    \caption{\raggedright MVNet}
    \label{Isabel_P_MVNet}
\end{subfigure}
~
\begin{subfigure}[t]{0.24\linewidth}
    \centering
    \includegraphics[width=\linewidth]{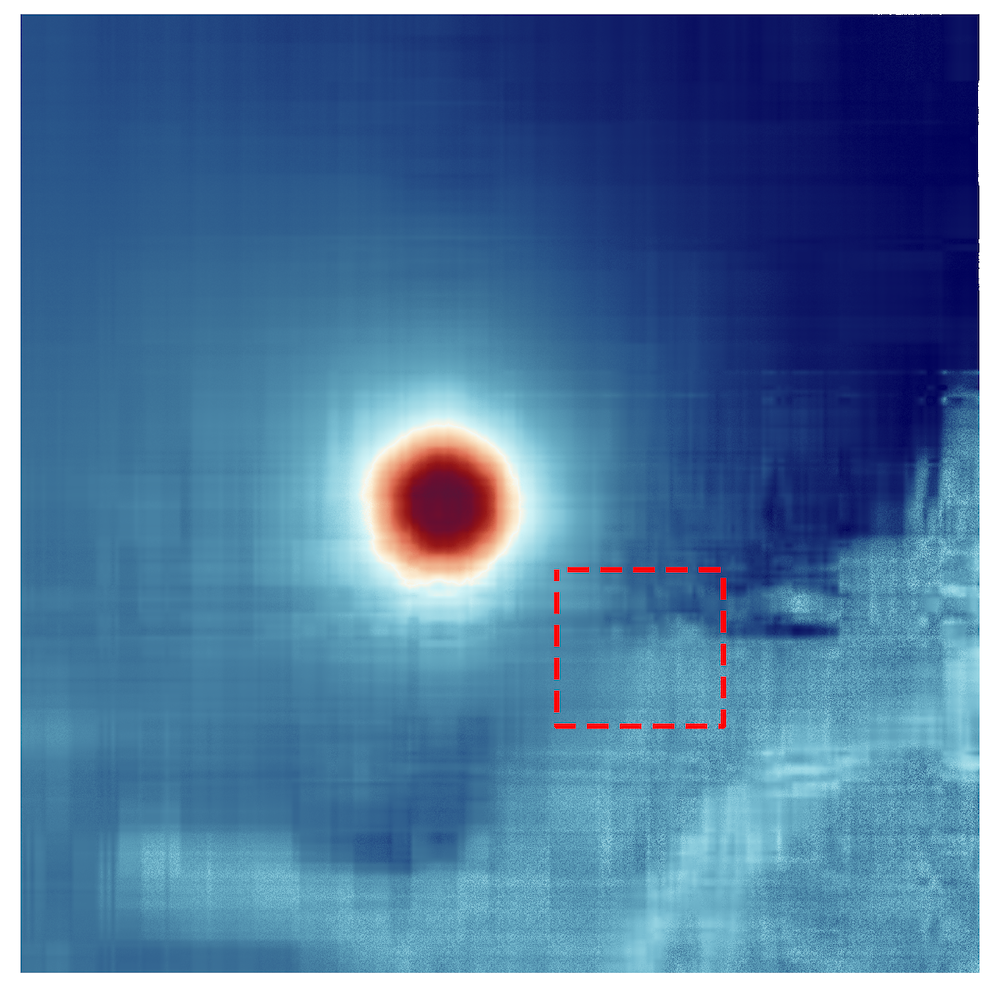}
    \caption{\raggedright TTHRESH}
    \label{Isabel_P_TTHRESH}
\end{subfigure}
~
\begin{subfigure}[t]{0.24\linewidth}
    \centering
    \includegraphics[width=\linewidth]{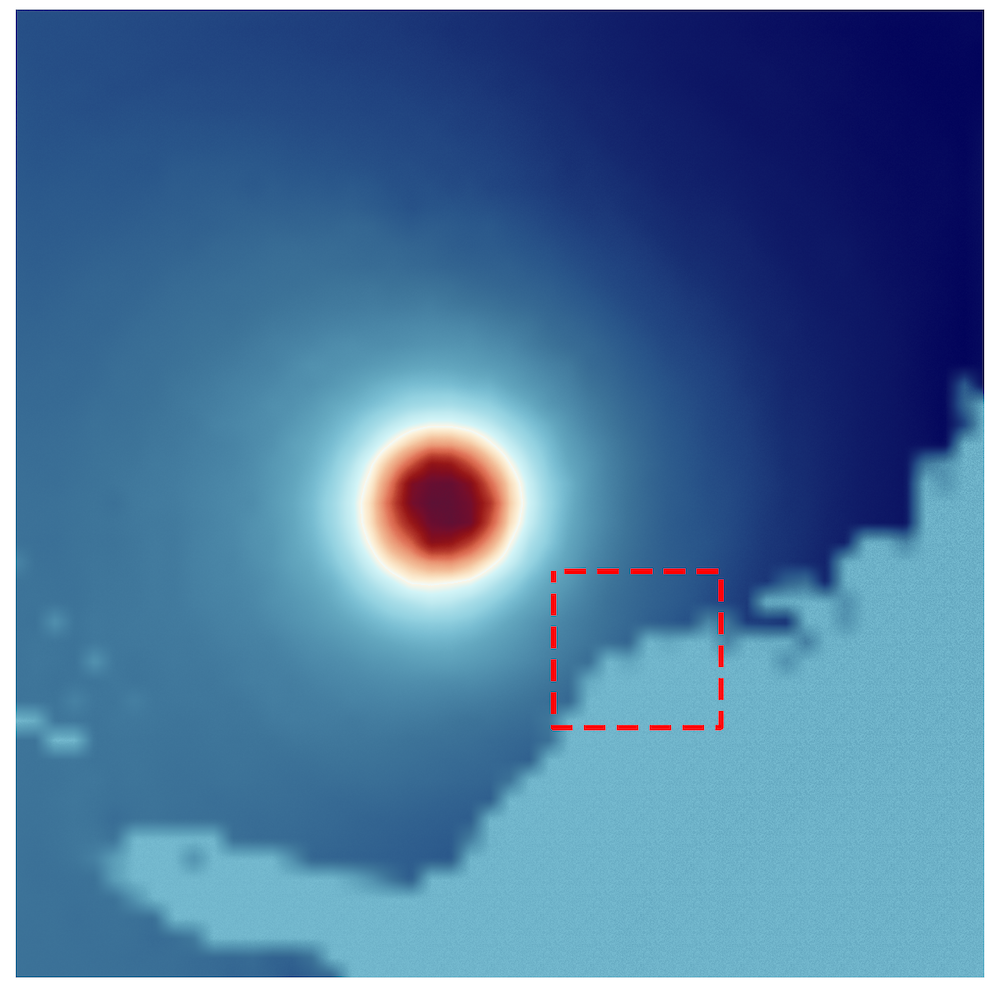}
    \caption{\raggedright LERP}
    \label{Isabel_P_LERP}
\end{subfigure}
\caption{Reconstructed data visualization for Pressure (P) variable of Isabel dataset. Images generated by ground truth (GT) data, MVNet, TTHRESH, and LERP are shown from left to right. We observe that the generated image is the most similar to the ground truth, while the TTHRESH and LERP-generated images have artifacts (as shown by the red dotted box).}
\label{isabel_vis_comp}
\end{figure*}

\begin{figure*}[thb]
\centering
\begin{subfigure}[t]{0.24\linewidth}
    \centering
    \includegraphics[width=\linewidth]{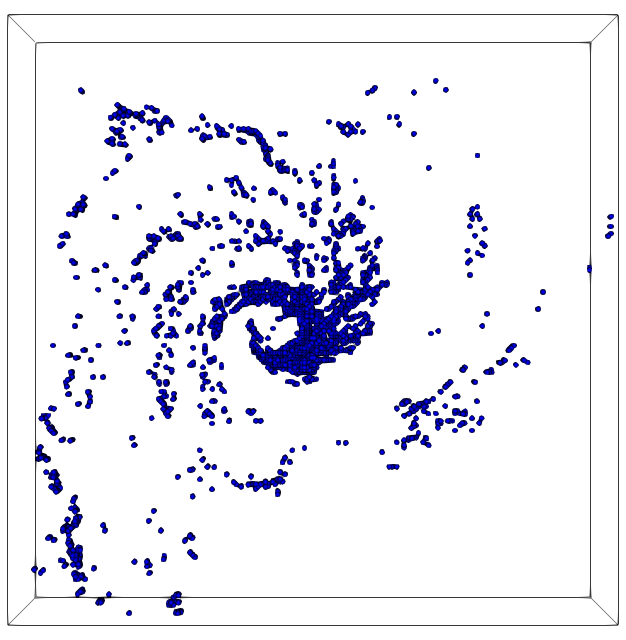}
    \caption{QDV result for ground truth.}
    \label{isabel_actual_qdv}
\end{subfigure}
~
\begin{subfigure}[t]{0.24\linewidth}
    \centering
     \includegraphics[width=\linewidth]{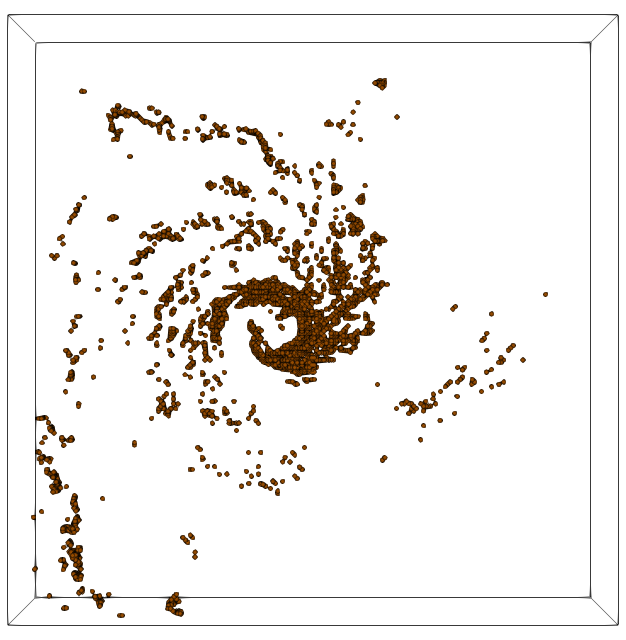}
    \caption{QDV result for MVNet.}
    \label{isabel_mvnet_qdv}
\end{subfigure}
~
\begin{subfigure}[t]{0.24\linewidth}
    \centering
    \includegraphics[width=\linewidth]{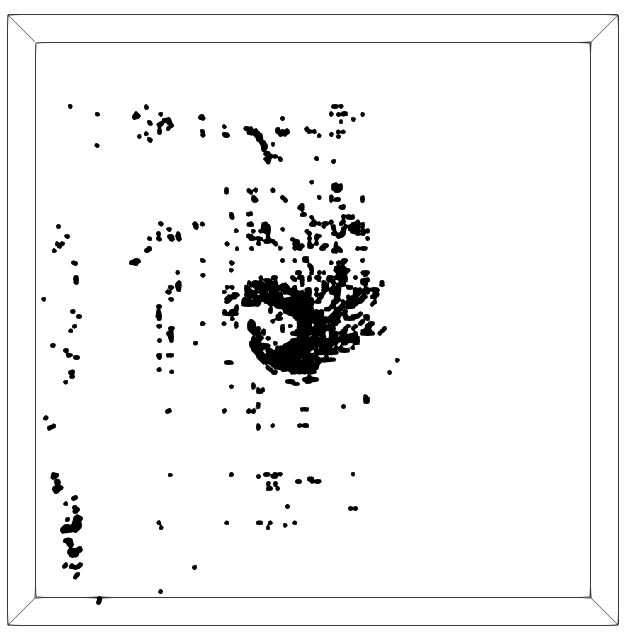}
    \caption{QDV result for TTHRESH.}
    \label{isabel_tthresh_qdv}
\end{subfigure}
~
\begin{subfigure}[t]{0.24\linewidth}
    \centering
    \includegraphics[width=\linewidth]{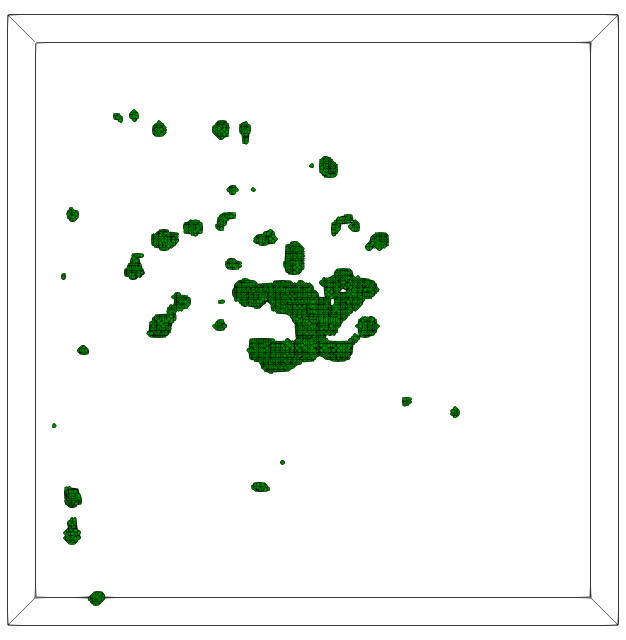}
    \caption{QDV result for LERP.}
    \label{isabel_lerp_qdv}
\end{subfigure}
\caption{Query-driven visualization for Hurricane Isabel dataset. A multivariate query: (CLOUD \textgreater 0.0001 AND \textless 0.002) AND (PRECIPITATION \textgreater 0.0001 AND \textless 0.0093) AND (QVAPOR \textgreater 0.01 AND \textless 0.0235) is shown. We observe that MVNet produces the most accurate result among the three methods.}
\label{qdv_isabel}
\end{figure*}

\begin{figure*}[thb]
\centering
\begin{subfigure}[t]{0.24\linewidth}
    \centering
    \includegraphics[width=\linewidth]{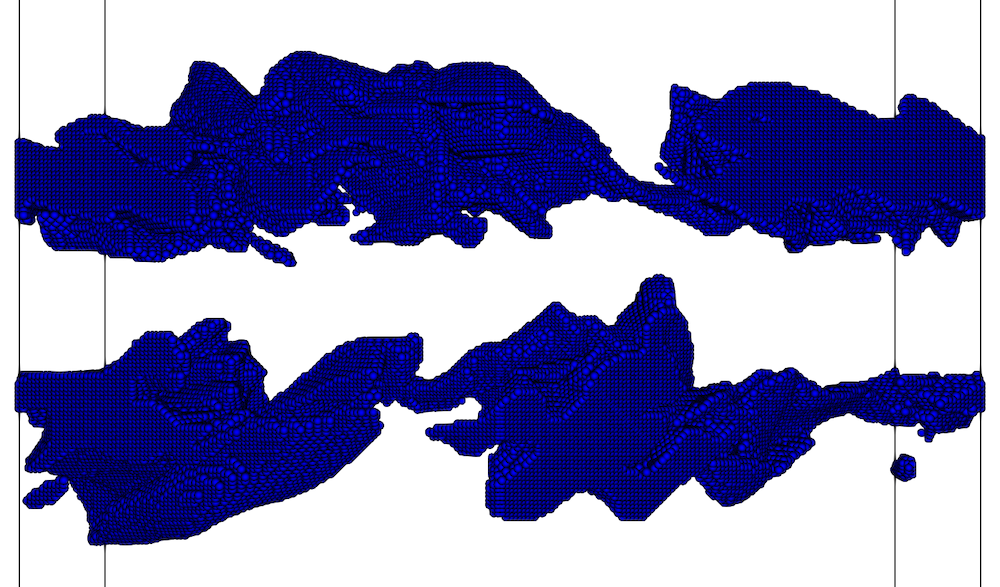}
    \caption{\raggedright QDV result for ground truth.}
    \label{comb_gt_qdv}
\end{subfigure}
~
\begin{subfigure}[t]{0.24\linewidth}
    \centering
     \includegraphics[width=\linewidth]{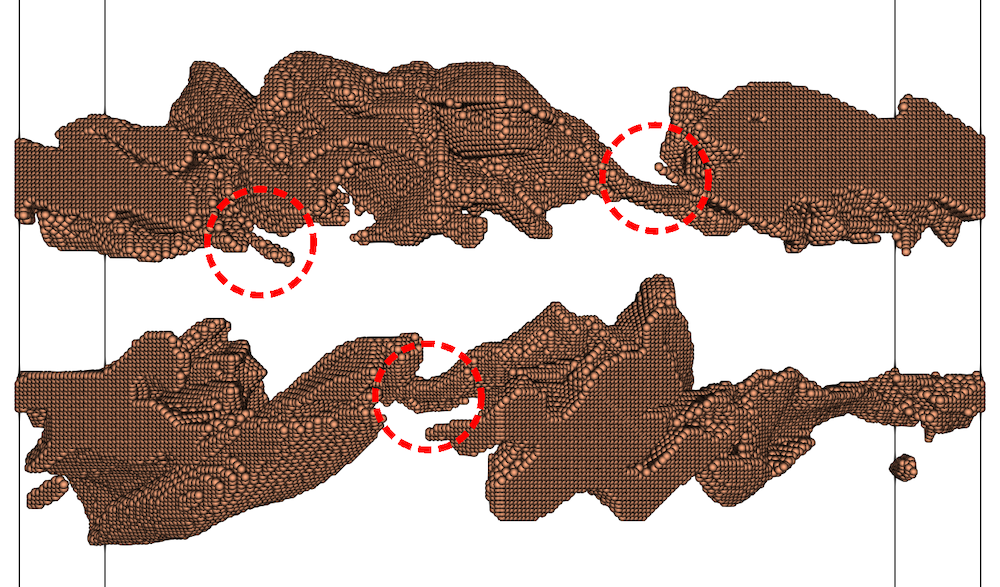}
    \caption{QDV result of MVNet.}
    \label{comb_mvnet_qdv}
\end{subfigure}
~
\begin{subfigure}[t]{0.24\linewidth}
    \centering
    \includegraphics[width=\linewidth]{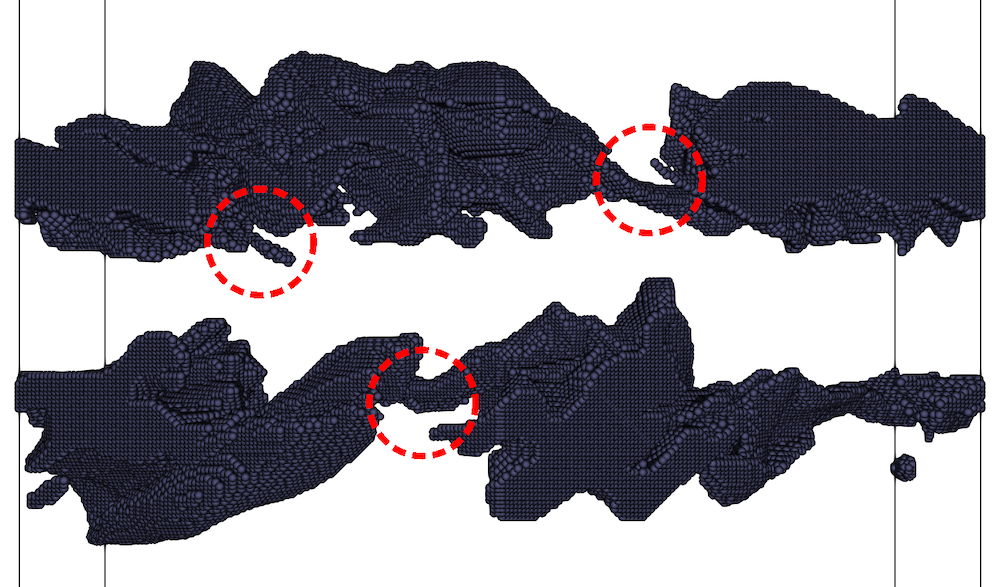}
    \caption{QDV result of TTHRESH.}
    \label{comb_tthresh_qdv}
\end{subfigure}
~
\begin{subfigure}[t]{0.24\linewidth}
    \centering
    \includegraphics[width=\linewidth]{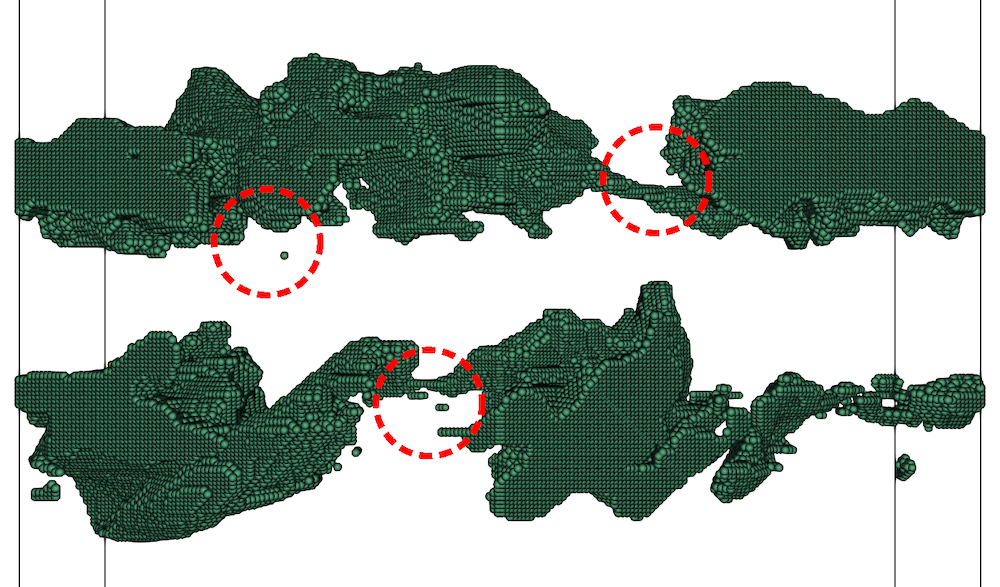}
    \caption{QDV result of LERP.}
    \label{comb_lerp_qdv}
\end{subfigure}
\caption{Query-driven visualization for the Combustion dataset. A multivariate query: (Mixfrac \textgreater 0.3 AND \textless 0.7) AND (Y\_OH \textgreater 0.006 AND \textless 0.1) is shown. We observe that while MVNet produces the most accurate result quantitatively, visually, MVNet and TTHRESH produce similar results, whereas LERP is the least accurate.}
\label{qdv_comb}
\end{figure*}

\begin{figure*}[!thb]
\centering
\begin{subfigure}[t]{0.24\linewidth}
    \centering
    \includegraphics[width=\linewidth]{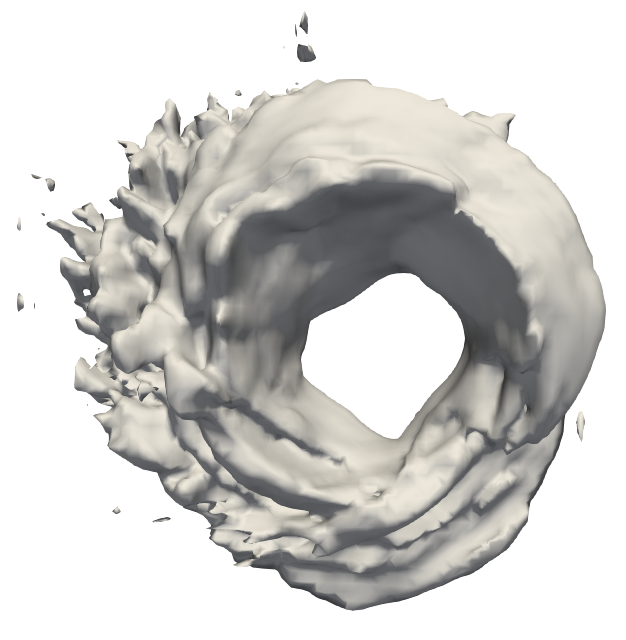}
    \caption{GT}
    \label{isabel_isosurf_41_GT}
\end{subfigure}
~
\begin{subfigure}[t]{0.24\linewidth}
    \centering
    \includegraphics[width=\linewidth]{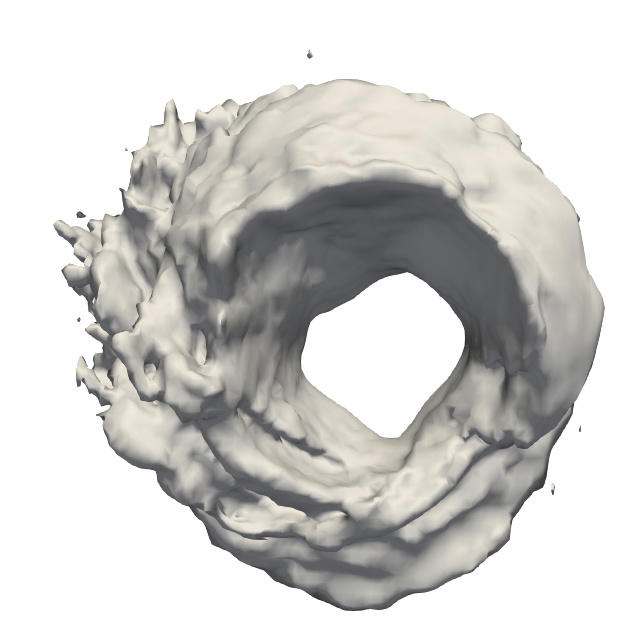}
    \caption{MVNet}
    \label{isabel_isosurf_41_mvnet}
\end{subfigure}
~
\begin{subfigure}[t]{0.24\linewidth}
    \centering
    \includegraphics[width=\linewidth]{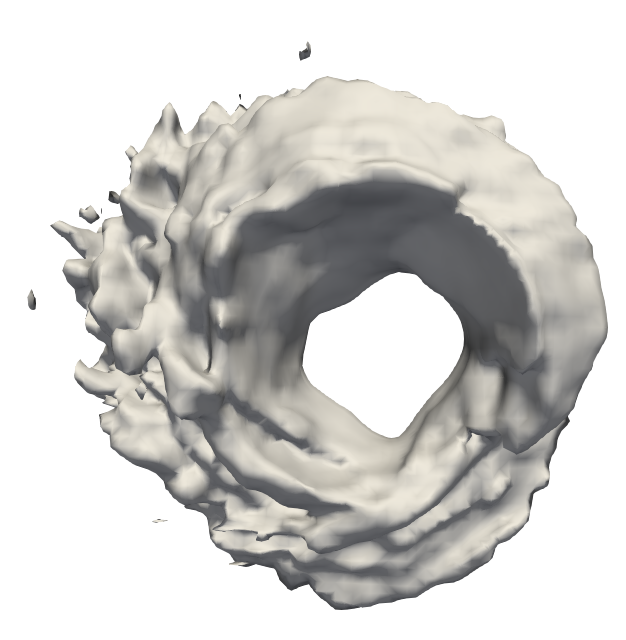}
      \caption{TTHRESH}
    \label{isabel_isosurf_41_tthresh}
\end{subfigure}
~
\begin{subfigure}[t]{0.24\linewidth}
    \centering
    \includegraphics[width=\linewidth]{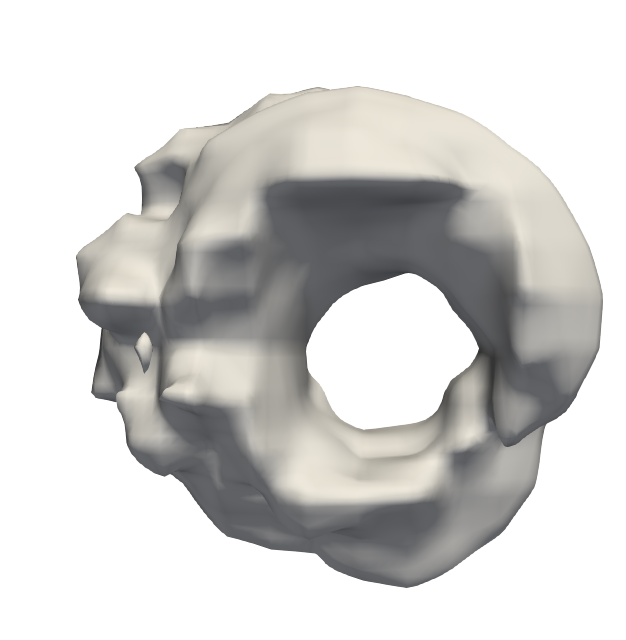}
    \caption{LERP}
    \label{isabel_isosurf_41_lerp}
\end{subfigure}
\caption{visualization of isosurface of Velocity Magnitude at isovalue=41 of Isabel dataset with ground truth (GT) shown in Fig.~\ref{isabel_isosurf_41_GT}. Isosurface produced by LERP is the least accurate, while MVNet and TTHRESH produce visually comparable results.}
\label{isabel_isosurf_vis}
\end{figure*}

\end{document}